Julie Greensmith · Jan Feyereisl · Uwe Aickelin

# The DCA: SOMe Comparison

A comparative study between two biologically-inspired algorithms

**Abstract** The Dendritic Cell Algorithm (DCA) is an immune-inspired algorithm, developed for the purpose of anomaly detection. The algorithm performs multi-sensor data fusion and correlation which results in a 'context aware' detection system. Previous applications of the DCA have included the detection of potentially malicious port scanning activity, where it has produced high rates of true positives and low rates of false positives. In this work we aim to compare the performance of the DCA and of a Self-Organizing Map (SOM) when applied to the detection of SYN port scans, through experimental analysis. A SOM is an ideal candidate for comparison as it shares similarities with the DCA in terms of the data fusion method employed. It is shown that the results of the two systems are comparable, and both produce false positives for the same processes. This shows that the DCA can produce anomaly detection results to the same standard as an established technique.

## 1 Introduction

The Dendritic Cell Algorithm (DCA) is an immune-inspired algorithm, which is the latest addition to a family of algorithms termed Artificial Immune Systems (AIS). Such systems use abstract models of particular components of the human immune system to produce systems which perform similar computational tasks to those seen in the human body. Previous approaches drew inspiration from the adaptive immune system producing algorithms such as clonal selection [14] and negative selection [33]. The negative selection approach has been used extensively in the domain of computer security specifically in the detection of potential intrusions [44]. Such systems enjoyed initial success, but have since been plagued with problems regarding scalability and the generation of excessive numbers of false positive alerts.

Aickelin *et al.* [1] proposed an alternative which suggests that successful AIS should be constructed using the 'danger theory' as inspiration. The danger theory suggests that the immune system responds to signals generated by the host cells (i.e. by the tissue) during cell stress, ultimately leading to the targeting of proteins present under the conditions of cell stress. It is a competing immunological theory, though it is still widely debated within immunology itself. This theory is centred around the detection of 'danger signals', which are released as a result of unplanned cell death, a process termed *necrosis*. Upon the detection of danger signals, the immune system can be activated. In the absence of necrosis, cells may die in a controlled manner as part of the regulatory processes found within the tissue, in a

Julie Greensmith, Jan Feyereisl and Uwe Aickelin
School of Computer Science
University of Nottingham
Wollaton Road
Nottingham, NG8 1BB, UK
E-mail: [jqg, jqf, uxa]@cs.nott.ac.uk

process called *apoptosis*. Dendritic cells (DCs) are sensitive to both signal types and have the ability to stimulate or suppress the adaptive immune system. DCs are the intrusion detection agents of the human immune system, policing the tissue for potential sources of damage in the form of signals and for potential culprits responsible for the damage in the form of 'antigen'. Antigens are proteins which can be 'presented' to the adaptive immune system by DCs, and can belong to pathogens or to the host itself.

The DCA incorporates danger-based DC biology to form an algorithm that is both truly bio-inspired and is capable of performing anomaly detection. It is a population based algorithm, where multiple DCs are programmed to process signals and antigen. 'Signals' are mapped to context information, such as the behaviour of a monitored, e.g. CPU usage or network traffic statistics. 'Antigens' are mapped as potential causes for the changes in behaviour, e.g. the system calls of a running program. The DCA correlates the antigen and signal information across the population of cells to produce a consensus coefficient value which is assessed to determine anomalous antigen.

The DCA has been successfully applied to a subset of intrusion detection problems, focussing on port scan detection. Port scans are used to establish network layout and to uncover vulnerable computers. The detection of the scanning phase of an attack can be highly beneficial, as upon its detection the level of security can be increased across a network in response. The DCA has been applied to both ping scans and SYN scans in realtime and offline [27] [25]. The algorithm produced high rates of true positives and low rates of false positives.

While the performance of the DCA on these problems appears to be good, thus far no direct comparison has been performed with another system on the same port scan data. The need for a rigorous comparison is necessary to truly demonstrate the capabilities of this algorithm. The signal processing component housed within the DCA bears some resemblance to the function of a neural network [13]. Given these superficial similarities, the obvious next-step for the development of the DCA is to compare its performance to that of a neural network based system, such as a Self-Organizing Map (SOM) [47].

SOM is a clustering method of unsupervised learning where high dimensional data is mapped to a lower dimensional space to create a feature map. This map is constructed from training data and consists of a series of interconnected nodes. Upon the analysis of the test data, the incoming data items are matched against nodes in the map with similar characteristics. SOM uses a similar process to a single-layer neural network to generate the map, and a simple distance metric is used to match the incoming test data to the most appropriate node. This technique can be used for anomaly detection as the training data can consist of normal data items, with unclustered data representing a potential anomaly. SOM is an excellent choice for comparison as it has a history of application within computer security and can be manipulated to use similar input data as used with the DCA.

The aim of this paper is to compare the DCA with a SOM. To achieve this aim the two algorithms are applied to the detection of an outbound SYN-based port scan using data captured from previous real-time experiments performed with the DCA [25]. The results of this comparison indicate that the DCA and SOM are both equally as effective at detecting SYN based port scans, and appear to make similar false positive errors. As a baseline a k-means clustering algorithm is applied to the signals in isolation.

This paper is structured as follows. In Section 2, the relevant background and context information is given regarding problems in computer security and how these problems relate to port scanning in addition to a summary of current port scanning techniques. In Sections 3 and 4, descriptions are given of the DCA and SOM respectively, including details of their implementations. In Section 5, the two approaches are compared experimentally. In Section 6 we perform an analysis and comparison between the two systems based on the obtained results and debate their differences and similarities, further validated by a baseline series. In the final sections we discuss the results of these comparisons and present the implications for the future of the DCA.

## 2 Related Work

### 2.1 Overview

As this paper encompasses a variety of techniques and concepts, this section is subdivided into three parts. Firstly, the problems associated with port scans are described followed by a description of current scan detection techniques. This is followed by the related computer security work in AIS, including the development of the DCA and the motivation for its development. This section continues with a brief overview of the use of various SOM algorithms in computer security.

## 2.2 Port Scanning and Detection

### 2.2.1 Introduction to Port Scanning

Insider attacks are one of the most costly and dangerous attacks performed against computerised systems, with a large amount of known intrusions of intrusions attributed to internal attackers [6]. This type of intrusion is defined through the attacker being a legitimate user of a system who behaves in an unauthorised manner. Such insider attacks have the potential to cause major disruption given that a large number of networks do not employ *internal firewalling* with many security countermeasures focussing on the detection of external intruders. Insiders frequently know and have access to network topology information. As insiders operate from within an organisation, this provides them with scope to abuse a weak link in the security chain, namely the end users. Having knowledge and relationships with other network users brings with it the potential to coerce passwords from legitimate users for the purpose of gaining access across multiple machines on a network. This information can be used to steal sensitive data, to cause damage to the network or to disguise the identity of the true attacker.

Such attacks frequently involve multiple stages. The initial stage is the information gathering stage, which is followed by monitoring of potential victims, finally involving an intrusion. The information gathering stage involves scouting the network for potential victim host machines to suit the nature of the attack. For example, the insider may wish to exploit an FTP service and would therefore search the network for hosts running FTP. Port scanning is used in the information gathering stage to retrieve which hosts are currently running on the network, the IP addresses and DNS names of each host, which 'ports' are currently 'open' and named running host services.

It is wise for an attacker to understand the network in question, to avoid wasting time trying to exploit machines which are not receptive to an attack. It is pointless attempting to attack a host which is no longer connected to the network! While scanners are not an 'intrusion' in the classical sense, they are often a pre-cursor to an actual attack, and evidence of sufficient scanning across a network can suggest that an attack may soon follow [39].

A port is a specific endpoint on a network, which is a virtual address as part of a virtual circuit. It is important to note that a port in this context is an abstract concept, not to be confused with a physical port such as a serial port. Ports allow for the direct exchange of information between two hosts. It is similar to a telephone number and is more specific than an IP address as it provides a direct connection between two endpoints. Probing a port with a packet leads to information on the state of the port and its host. Ports can be in three states if the scanned host is available, namely open, closed or filtered. Port scanning tools such as nmap [17] can be used to send packets to various ports on remote hosts to gain understanding regarding the status of the scanned host. The type of packet used to perform such probes can be one of a number of types, including Internet Control Message Protocol (ICMP) ping, TCP, and UDP. According to Bailey-Lee *et al.* [4], TCP SYN scans are the most commonly observed scan, accounting for over half of all scans performed.

Additionally the scans themselves can be performed in a number of different ways, varying the number of hosts scanned, the number of ports scanned, the IP address of the sender and the rate at which packets are sent. The combination of the number of hosts (IP addresses) scanned and the number of ports gives rise to three combinations of scan:

- *Horizontal Scans*: A wide range of IP addresses are scanned, though only one port per host is probed. This kind of scan can be used if a particular service is to be exploited, such as an FTP exploit whose success would depend on finding a host running the particular version of an FTP server/client.
- *Vertical Scans*: A range of ports are specified for a single IP address. This is a scan of a single host, where multiple services can be searched for. Example usage of this type of scan includes characterising a specific target such as a web server to assess if any exploitable services are currently running.

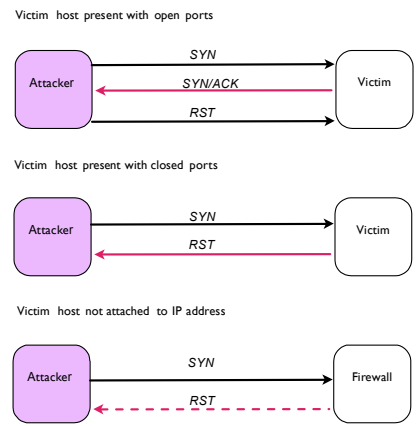

Fig. 1 The TCP/IP packet flow for SYN scans under various conditions.

– *Block Scans*: This is a combination of both types of scan, where multiple ports are probed across a range of IP addresses. This search is especially good for uncovering the topology of a network, for locating servers, printers, client hosts etc.

A SYN scan, apart from being the most prevalent scan, can be used to retrieve a substantial amount of information regarding the status of a host or a network. SYN scans exploit the '3-way handshake' employed by the TCP protocol. The 3-way handshake describes the specific manner in which two endpoints communicate with each other and is the initial step in opening a TCP data stream. For example two hosts are connected to one another via a switched ethernet, host A and host B. To initiate TCP connections between A and B, A sends a TCP SYN (synchronisation) packet to B. Provided that B can receive such TCP packets, B sends a SYN/ACK (synchronisation/acknowledgement) packet to A in response to the initial SYN request. Host A then replies with ACK packets which are sent until all data in the transmission stream is transferred. Upon completion, A sends B a FIN (finish) packet, signalling that the data transfer is complete. Port scans based on TCP SYN packets exploit a flaw in this 3-way connection method.

Instead of completing the 3-way handshake of 'SYN, SYN/ACK, ACK', upon receipt of the SYN/ACK packet, the scanning host does not send an ACK packet in response: depending on the initial response of remote host B depends on how the scanning host behaves. The various responses to SYN requests are depicted in Figure 1.

In the case of a host with an open port, the victim sends the attacker a SYN/ACK packet in response to the original SYN request. Instead of the attacker responding back with an ACK packet, a RST packet is sent instead. This leaves the TCP connection 'half open'. As the conversation remains incomplete, it does not appear in system logs, making the SYN scan stealthy. Should the scanned victim host have no open ports the attacker receives a RST packet. This however, informs the attacker that a host is connected to an IP address. If no host is associated with a scanned IP address, the connection will time out on the attackers machine, or the attacker will receive a "destination unreachable error" from an interim router or firewall.

### 2.2.2 Port Scan Detection

Port scan detection components are frequently integrated with commercial intrusion detection systems such as Snort [62]. The detection method used involves generating an 'alert' upon receiving 'X' connections within 'Y' seconds from host IP address 'Z'. There are three problems with this method. Firstly, the majority of port scanners are equipped with the facility to fake the IP address, often called spoofing, of the source machine, rendering Z unreliable. Secondly most scanners allow for the randomisation of the duration between the sending of individual packets, including slowing the rate of sending to avoid detection. This makes the detection of the scan difficult if Y is defined as a constant value. Finally, should the port scan detector be host based, the performance of a detector in response to an incoming scan will be impaired for the detection of horizontal scans, as only a single connection is made to each host. This problem can be overcome through the detection of outbound port scans, as per extrusion detection [6].

To overcome some of these problems a handful of systems are developed as dedicated port scan detectors. For example 'Spice' developed by Staniford *et al.* [68] incorporates an anomaly probability score to dynamically adjust the duration of Y . This is useful for the detection of stealthier scans which use randomised or slowed rates of scan packets sending. This detection technique is termed *reverse sequential hypothesis testing*. It is used to further extent in research performed by Jung *et al.* [39], where it is combined with a network based approach and is used to identify potentially malicious remote hosts in addition to detecting scanning activity.

In our previous research, the DCA is applied to the detection of various port scans. The DCA is implemented as a host based system monitor, detecting the performance of an outbound port scan, in attempt overcome some of the problems with using static time windows. Initially the DCA is applied to the detection of simple ICMP ping scans [29], where the algorithm was used in real-time and produced high rates of true positives and low rates of false positives.

In addition, DCA is used in the detection of a standard SYN scans, also in a real-time environment [27]. The results of this study show that the DCA shows promise as a successful port scan detector. However the results presented were preliminary and as the experiments were performed 'live' in real-time, certain sensitivity could not be performed. Therefore, it is necessary to take this investigation further and explore this application with more rigour. The experiments described in this paper are extension experiments from the SYN scan data used in Greensmith and Aickelin [25].

## 2.3 Artificial Immune Systems and Security

### 2.3.1 Immunity by Design

Numerous computer security approaches are based on the principles of *anomaly detection.* This technique involves producing a profile of normal host behaviour, with any significant deviation from this profile presumed to be malicious or anomalous. Various AIS have been applied as anomaly detection algorithms within the field of computer security, given the obvious parallel of fighting computer viruses with a computer immune system [18]. The research of AIS in security has extended past the detection of viruses and has focussed on network intrusion detection [45].

The AIS algorithms used in security are generally based on the principles of "self-nonself discrimination". This is an immunological concept that the body has the ability to discriminate between self and nonself (i.e. foreign) protein molecules, termed *antigen.* The natural mechanism by which the body learns this discrimination is termed *negative selection*. In this process, self-reactive immune cells are deleted during a 'training period' in embryonic development and infancy. This results in a tuned population of cells, poised to react against any threat which is deemed nonself. These principles are used to underpin the supervised negative selection algorithm. Negative selection itself is described eloquently in a number of sources including the work of Hofmeyr and Forrest [32], Ji and Dasgupta [37] and Balthrop *et al.* [5].

Following its initial success in the detection of system call anomalies [32], the negative selection approach was applied to the detection of anomalous network connections [44], where potential problems with scaling and excessive false positive rates were uncovered. These empirical studies suggest that negative selection might not be a suitable algorithm for use in computer security, with these notions confirmed by the theoretical work performed by Stibor *et al.* [69]. Further analysis performed by the same authors has given insights into the theoretical reasons for negative selection's problems [70], with more evidence presented recently by Stibor *et al.* in [71].

### 2.3.2 The Danger Project: The Missing Link?

The criticisms of negative selection have to some degree overwhelmed the positive aspects of its development and have in some respect discredited the use of AIS in security [66]. Issues such as scaling cannot be ignored, especially as anomaly detection is often required to be performed in real-time. The theoretical proofs of these problems are evident and duly noted by the wider AIS community [72].The question of how to overcome these problems remains at the forefront of AIS research, focussing on the incorporation of more advanced immunology.

An interdisciplinary approach is presented by Aickelin *et al.* [1], developed in 2003 through the Danger Project. Aickelin *et al.* believe that some of the problems shown with negative selection approaches can be attributed to its biological naivety. It is recognised that the negative selection algorithm is based on a naive model of central tolerance developed in the 1950s [12].

Aickelin *et al.* propose that through close collaboration with immunologists, computer scientists will be able to develop more biologically realistic AIS which could potentially overcome the problems of false positives and scaling observed with negative selection [1]. The DCA is developed using this interdisciplinary approach [26], drawing inspiration from DCs as it is now widely accepted that these cells are a major control unit in the human immune system.

The Danger Project brought innate immunology in to the AIS spotlight, as the innate immune system is shown as responsible for the initial pathogen detection [52]. From this emerged two streams of research which were based on innate principles, the Dendritic Cell Algorithm and the `libtissue` system and its related algorithms [73]. The DCA will be explained in detail in Section 3, and is based on an abstract model of the behaviour of natural DCs. The *libtissue system* is an innate immune framework implemented as an API (application programming interface) [75].

### 2.3.3 Summary

AIS have been used within computer security for over a decade. Despite its initial success, the negative selection algorithm was not as useful as first thought due to problems with scaling and the generation of excessive amounts of false positives. These negative aspects have been shown both theoretically and experimentally. To remedy this problem, the research proposed by Aickelin *et al.* [1] developed into the DCA and the `libtissue` framework, both of which have shown promise in the areas of port scan and exploit detection respectively.

## 2.4 SOM and Security

The Self-Organizing Map algorithm was developed by Teuvo Kohonen more than two decades ago [46], yet its success in various fields of science, over the years, surpasses many other neural inspired algorithms to date. The algorithm's strengths lie in a number of important scientific domains. Namely visualisation, clustering, data processing, reduction and classification. In more specific terms SOM is an unsupervised learning algorithm that is based on the competitive learning mechanism with self-organizing properties. Besides its clustering properties SOM can also be classed as a method for multidimensional scaling and projection.

SOM algorithms have been first applied to computer security applications almost ten years after the algorithm's inception [19]. The majority of existing research however is limited to anomaly detection, particularly network based intrusion detection [16]. Some work has been done on host based anomaly detection using Kohonen's algorithm, however such work is rare [35], which is surprising, due to the algorithms suitability to handle multidimensional, thus multi-signal data. On numerous occasions SOM algorithms have been used as a pre-processor to other computational intelligence tools, such as Hidden Markov Models (HMM) [11] [10] [43] or Radial Basis Function Networks [36]. Comparisons of SOM algorithms with other anomaly detection approaches have been performed on numerous occasions in the past. Notably a comparison with HMM [76], AIS [21] [22], traditional neural networks [63] [50] [38] [49] [8], k-means clustering [42] as well as Adaptive Resonance Theory [3].

### 2.4.1 SOM in Intrusion Detection

Besides the above mentioned comparisons of SOM algorithms with various computational intelligence techniques, here we will describe the use of SOM algorithms in areas related to our comparison or that could be of interest to the general reader.

A seminal paper on the use of SOM algorithms for intrusion detection was presented by Ramadas *et al.* [59]. Their work employed the original SOM algorithm as a network based anomaly detection module for a larger IDS. Besides being able to monitor all types of network traffic including SMTP protocol (email), the authors state that the SOM algorithm is particularly suitable to detect buffer-overflow attacks. However, as with the majority of anomaly detection systems, the algorithm struggles to recognise attacks which resemble normal behaviour in addition to boundary case behaviour, giving rise to false positives.

Buffer overflow [20] attack detection was also tackled by Rhodes [60] using a multilayer SOM, monitoring payloads. Bolzoni [9] also looked at payload monitoring using SOM by employing a two-tier architecture system.

Gonzalez and Dasgupta [21] compared SOM against another AIS algorithm. Their Real-Valued Negative Selection algorithm is based on the original Negative Selection algorithm proposed by Forrest *et al.* [18] with the difference of using a new representation. The original Negative Selection algorithm has been applied to Intrusion Detection problems in the past and has received some criticism regarding its "scaling problems" [44]. Gonzalez and Dasgupta argue that their new representation is the key to avoiding the scaling issues of the original algorithm. Their results showed that for their particular problem the SOM algorithm and their own were comparable overall.

Albarayak *et al.* [2] proposed a unique way of combining a number of existing SOM approaches together in a node based IDS. Their thesis is of automatically determining the most suitable SOM algorithm for each node within their system. Such a decision can be achieved using heuristic rules that determine a suitable SOM algorithm based on the nodes environment.

Miller and Inoue [55] suggested using multiple intelligent agents, each of which contains a SOM on its own. Such agents combine a signature and anomaly based detection technique in order to achieve a collaborative IDS, which is able to improve its detection capabilities with the use of reinforcement learning.

DeLooze [15] employed an ensemble of SOM networks for the purpose of an IDS as well as for attack characterisation. Genetic algorithms were used for attack type generation, subsequently employed as part of an IDS that is able to discriminate the type of attack that has occurred.

SOM algorithms have also been used for the analysis of executables. Yoo [78] analysed windows executables by creating maps of EXE files before and after an infection by a virus. Such maps have been subsequently analysed visually and found to have contained patterns, which can be thought of as virus masks. The author concludes by stating that such masks can be used in the future for virus detection in a similar manner to current anti-virus techniques. The difference being that a single mask could detect viruses from a whole virus family rather than being able to find only a single variant.

Besides the original single network SOM algorithm, the previously described newer SOM variations were employed for various security research scenarios. A number of papers discuss the advantages of using multiple or hierarchical SOM networks in contrast to a single network SOM. These include the work of Sarasamma *et al.* [64], Lichodzijewski *et al.* [51] and Kayacik *et al.* [41] [30] who all used various versions of the Hierarchical SOM or employed multiple SOM networks for the purpose of intrusion detection. Kayacik *et al.* [41] state that the best performance is achieved using a 2-layer SOM and that their results are by far the best of any unsupervised learning based IDS to date.

*2.4.2 Visualisation using SOM in IDS*

Due to the SOM algorithm's capability of visualising multidimensional data in a meaningful way, its use lends itself ideally to its application in visualising computer security problems. Gonzalez *et al.* [23] use this ability to visualise the self non-self space that they use for anomaly detection. This visualisation presents a clear discrimination of the different behaviours of the monitored system. Hoglund *et al.* [34] on the other hand employed visualisation of user behaviour. In their work various host based signals were used for monitoring of users. A visual representation was subsequently presented to administrators in order for them to be able to make an informed decision in case of unacceptable user behaviour.

# 3 The Dendritic Cell Algorithm

## 3.1 Natural DCs

The DCA is based on the observed functions of natural dendritic cells. DCs are natural intrusion detection agents, who monitor the host tissue for evidence of damage. In the human body, DCs have a dual role, as garbage collectors for tissue debris and as commanders of the adaptive immune system. DCs belong to the innate immune system, and do not have the adaptive capability of the lymphocytes of the adaptive immune system. DCs exist in three states of differentiation, *immature, semi-mature* and *mature*, which determines their exact function [52]. Modulations between the different states are dependent upon the receipt of signals while in the initial or immature state. Signals which indicate damage cause a transition from immature to mature. Those signals indicating good health in the monitored tissue cause a transition from immature to semi mature. The signals in question are derived from numerous sources, including pathogens, from healthy dying cells, from damaged cells and from inflammation. Each DC has the capability to combine the relative proportions of input signals to produce its own set of output signals. Input signals processed by DCs are categorised based on their origin:

PAMPs: Pathogenic associated molecular patterns are proteins expressed exclusively by bacteria, which can be detected by DCs and result in immune activation [56]. The presence of PAMPS usually indicates an anomalous situation.

Danger signals: Signals produced as a result of unplanned necrotic cell death. On damage to a cell, the chaotic breakdown of internal components form danger signals which accumulate in tissue [53]. DCs are sensitive to changes in danger signal concentration. The presence of danger signals may or may not indicate an anomalous situation, however the probability of an anomaly is higher than under normal circumstances.

Safe signals: Signals produced via the process of normal cell death, namely apoptosis. Cells must apoptose for regulatory reasons, and the tightly controlled process results in the release of various signals into the tissue [54]. These 'safe signals' result in immune suppression. The presence of safe signals almost certainly indicate that no anomalies are present.

Inflammation: Various immune-stimulating molecules can be released as a result of injury. Inflammatory signals and the process of inflammation is not enough to stimulate DCs alone, but can amplify the effects of the other three categories of signal [67]. It is not possible to say whether an anomaly is more or less likely if inflammatory signals are present. However, their presence amplifies the above three signals.

Dendritic cells act as natural data fusion agents, producing various output signals in response to the receipt of differing combinations of input signal. The relative concentration of output signal is used to determine the exact state of differentiation, expressed by the production of two molecules, namely the mature and semi-mature output signals. During this phase they are exposed to varying concentrations of the input signals. Exposure to PAMPs, danger signals and safe signals causes the increased production of *costimulatory molecules*, and a resulting removal from the tissue and migration to a local lymph node.

DCs translate the signal information received in the tissue into a context for antigen presentation, i.e. the antigen presented in an overall 'normal' or 'anomalous' context. The antigen collected while in the immature phase is expressed on the surface of the DC. Whilst in the lymph node, DCs seek out T-lymphocytes (T-cells) and attempt to bind expressed antigen with the T-cells variable region receptor. T-cells with a high enough affinity for the presented antigen are influenced by the output signals of the DC. DCs exposed to predominantly PAMPs and danger signals are termed 'mature DCs'; they produce mature output signals, which activate the bound T-cells. Conversely, if the DC is exposed to predominantly safe signals the cell is termed semi-mature and antigens are presented in a safe context, as little damage is evident when the antigen is collected. The balance between the signals is translated via the signal processing and correlation ability of these cells. The overall immune system response is based on the systemic maturation state average of the whole DC population on a per antigen basis. An abstract view of this process is presented in Figure 2.

## 3.2 Algorithm Overview

The purpose of the DCA is to correlate disparate data-streams in the form of antigen and signals and to label groups of identical antigen as 'normal' or 'anomalous'. This algorithm is population based with each 'cell' expressed as an 'agent'. The DCA is not only a classification algorithm, it also shares properties with certain filtering techniques. It provides information representing how anomalous a group of antigens are, not simply if a data item is anomalous or not. This is achieved through correlating a time-series of input signals with a group of antigen. The signals used are pre-normalised and pre-categorised data sources, which reflect the behaviour of the system being monitored. The co-occurrence of antigen and high/low signal values forms the basis of categorisation for the antigen data. The primary components of a DC based algorithm are as follows:

1. Individual DCs with the capability to perform multi-signal processing;
2. Antigen collection and presentation;
3. Sampling behaviour and state changes;
4. A population of DCs and their interactions with signals and antigen;
5. Incoming signals and antigen, with signals pre-categorised as PAMP, danger, safe or inflammation;
6. Multiple antigen presentation and analysis using 'types' of antigen;
7. Generation of anomaly coefficient for various different types of antigen.

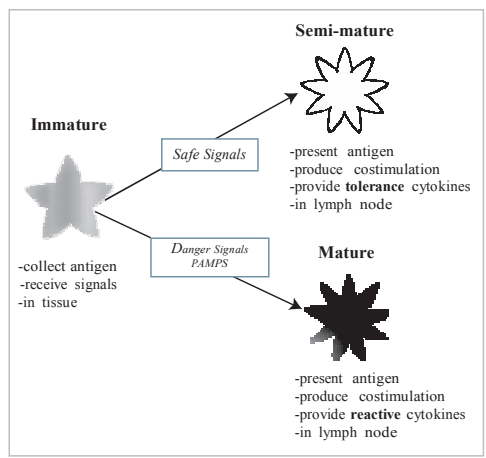

Fig. 2 An abstract view of DC maturation and signals required for differentiation where cytokines are molecular messengers between immune system cells.

Table 1 Table of cumulative output signals and their associated implications for the DCA.

| Output signal | Function |
|---|---|
| CSM signal | assessed against a threshold to limit the duration of DC signal and antigen sampling, based on a migration threshold |
| Semi-mature signal | terminal state to semi-mature if greater than resultant mature signal value |
| Mature signal | terminal state to mature if greater than resultant semi-mature signal value |

Whilst in the immature state, the DC has three functions, which are performed each time a single DC is updated, with the exact nature of this processing given in Section 3.4:

1. *Sample antigen:* the DC collects antigen from an external source (in our case, from the 'tissue') and places the antigen in its own antigen storage data structure.
2. *Update input signals:* the DC collects values of all input signals present in the signal storage area.
3. *Calculate interim output signals:* at each iteration each DC calculates three temporary output signal values from the received input signals, with the output values then added to form the cell's cumulative output signals.

The signal processing used to transform the input to interim output signals is shown in Figure 3, with the implications of each output signal given in Table 1. Costimulatory molecule (CSM) signal is used to limit the sampling duration of an individual cell. Each cell is assigned a migration threshold value upon creation. The CSM values are incremented each time a cell receives signal input. Once the CSM exceeds the cell's migration threshold the cell is removed from the sampling population for analysis. Different cells sample for different durations as each cell is assigned a random migration threshold value (within a range to match the normalised range of the input signals). This leads to a 'time-window' effect which adds robustness to the algorithm, as the migration is signal dependent with no fixed time dependency. During periods where signal values are large the rate of migration is higher and therefore tighter coupling is given to the signal and antigen data. This effect is explored in more detail in Oates *et al.* [58] where a theoretical analysis is provided.

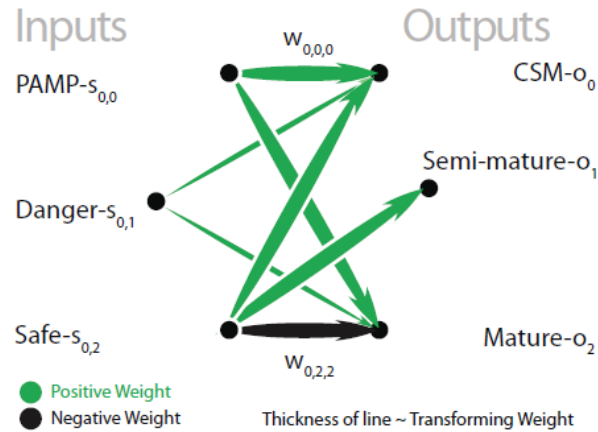

Fig. 3 A representation of the three calculations performed by each DC, per update cycle, to derive the cell's outputs through fusing together the signal inputs. Two sample notations for the weights are shown. The inflammation signal (if any) applies to all transformations shown and is therefore not depicted. Details of the notation are given in Section 3.4.

Upon transition to a matured state, the DCs output signals are assessed to form the context of all antigen collected. A higher 'mature' output signal value results in the assignment of a context value of 1 to the DC, whereas a higher 'semi-mature' output signal value results in a context value of 0. All antigen sampled by the DC over its lifetime are output with the assigned context value. Upon completion of all data processing the mean context in which the antigens are presented is calculated deriving an anomaly coefficient value per antigen type. Each antigen (suspect data item) is not unique, but several identical antigens are sampled. The *mature context antigen value* - MCAV, is used to assess is a particular antigen type is anomalous. The derivation of the MCAV per antigen type is shown in Equation 1,

$$MCAV_x = \frac{Z_x}{Y_x} \qquad (1)$$

where $MCAV_x$ is the MCAV coefficient for antigen type x, $Z_x$ is the number of mature context antigen presentations for antigen type x and $Y_x$ is the total number of antigen presented for antigen type x.

The effectiveness of the MCAV is dependent upon the use of *antigen types*. This means that the input antigens are not unique in value, but belong to a population in themselves. For example, the ID value of a running program is used to form antigen, with each antigen generated every time the program sends an instruction to the low level system. Therefore a population of antigen is used, linked to the activity of the program and all bearing the same ID number.

To process signals, antigens and cells the DCA uses two virtual compartments: tissue and lymph nodes. The tissue is used as storage for antigens and signals and the lymph node is used for MCAV generation. The tissue consists of antigen and signal containers from which the DCs sample the input data. Signals are updated at regular intervals and are not removed upon sampling by a DC. Antigens are input in an event-driven manner and are removed from the tissue antigen store upon sampling by a DC.

We suggest that the updates of antigen, signals and cells are performed independently, based on previous experience with this algorithm. This is represented in Figure 4. The three updates need not occur simultaneously: this temporal correlation between asynchronously arriving data is performed by the processing of the cells themselves. The population dynamics are used to perform the actual anomaly detection. The ultimate classification of a particular type of antigen is derived not from a single DC, but from an aggregate analysis produced across the DC population over the duration of an experiment.

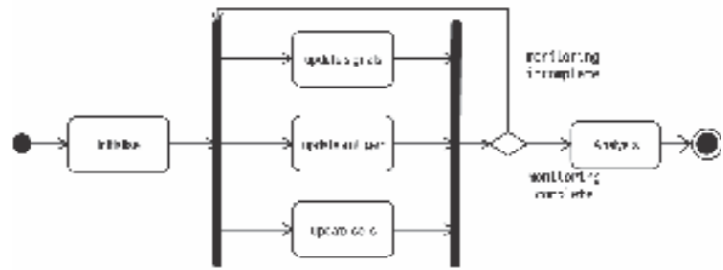

Fig. 4 A UML overview of the processes at the tissue level of the program, showing the asynchronous update of cells, signals and antigen. It also shows the two main stages of update and initialisation and subsequent analysis.

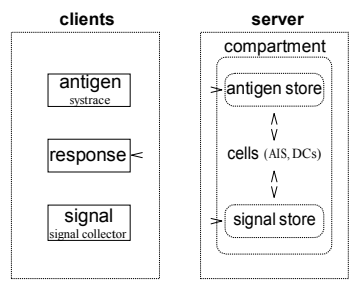

Fig. 5 Architecture used to support the DCA. Input data is processes via signal and antigen clients [75]. The algorithm utilises this data and resides on a server.

### 3.3 `libtissue`

The Danger Project [1] has produced a variety of research outcomes alongside the DCA. Such outcomes include the development of danger theory and DC based immunology [77]; a framework for developing immune inspired algorithms called `libtissue` [75]; an investigation into the interactions between the innate and adaptive immune system; artificial tissue [7] and the application of a naive version of the DCA for the security of sensor networks. `libtissue` is the API used within the Danger Project for the testing of ideas and algorithms, as shown in the works of Twycross [73] [75] and Greensmith *et al.* [28] [29].

`libtissue` is a library implemented in C which assists the development of immune inspired algorithms on real-world data. It is based on principles of innate immunology [74] [75], through the use of techniques from modelling, simulation and artificial life. It allows researchers to implement algorithms as a collection of cells, antigen and signals, interacting within a specified compartment. The implementation has a client/server architecture which separates data collection using clients, from data processing on a server, as shown in Figure 5.

Input data is processed using `libtissue` clients, which transform raw data into antigen and signals. Algorithms can be implemented within the `libtissue` server, as it provides all the required components such as the ability to define different cell types, specifying receptors, compartments and internal signals. Antigen and signal sources can be added to `libtissue` servers, facilitating the testing of the same algorithm with a number of different data sources. Input data from the client are passed to and represented in a compartment contained on a server known as the tissue compartment. This is a space in which cells, signals and antigen interact. Each tissue compartment has a fixed-size antigen store where collected antigens are placed. The tissue compartment also stores levels of signals, set by the input clients.

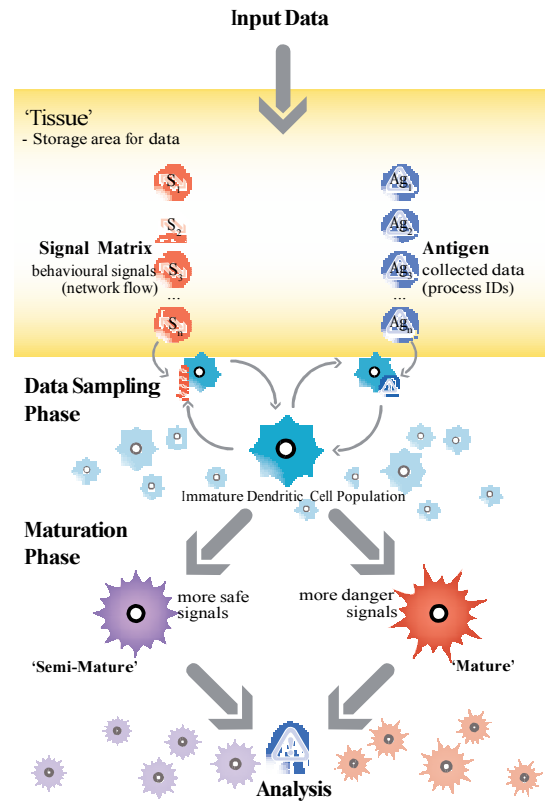

Fig. 6 Illustration of the DCA showing data input, continuous sampling, the maturation process and aggregate analysis.

### 3.4 DCA Implementation

The DCA is implemented as a `libtissue` server. Input signals are combined with antigen data, such as a program ID number. This is achieved through using the population of artificial DCs to perform aggregate sampling and data processing. Using multiple DCs means that *multiple* data items in the form of antigen are sampled *multiple times*. If a single DC presents incorrect information, it becomes inconsequential provided that the majority of the DCs population derive the correct cell context represented as the MCAV coefficient.

The DCA has two main stages: *initialisation*, and *update*. Initialisation involves setting various parameters and is followed by the update stage. The update stage can be decomposed into *tissue update* and *cell cycle*. Signal data is fed from the data-source to the tissue server through the tissue client and is updated at a user defined rate. The cell cycle involves the regular update of the DC population. Following the processing of all data, the MCAV coefficients are calculated for each antigen type. An overview of this is given in Figure 6.

The tissue update is a continuous process, whereby the values of the tissue data structures are refreshed. In this implementation, signals are updated at regular intervals - in our case, this is at a rate of once per second. The update of antigen occurs on an event-driven basis, with antigen items updated in the tissue each time new raw data appears in the system. The updated signals provide the input signals for the population of DCs.

The cell cycle is a discrete process occurring at a user defined rate of once per second in this research. Signals and antigen from the tissue data structures are accessed by the DCs during the cell cycle. This includes an update of every DC in the system with new signal values and antigen. The new values are processed and accumulated as the output vector of signals is generated. The cell cycle and update of tissue continues until a stopping criterion is reached. Finally, the aggregation stage is initiated, where all collected antigens are subsequently analysed and the MCAV per antigen derived. This procedure forms the required post-processing for use with this algorithm.

## 3.5 Parameters and Structures

The algorithm is described formally using the following terms.

– Indices:
   $i = 0, ..., I$ input signal index;
   $j = 0, ..., J$ input signal category index;
   $k = 0, ..., K$ tissue antigen index;
   $l = 0, ..., L$ DC cycle index;
   $m = 0, ..., M$ DC index;
   $n = 0, ..., N$ DC antigen index;
   $p = 0, ..., P$ DC output signal index.

– Parameters:
   $I$ = number of input signals per category (e.g. PAMP, Danger, Safe);
   $J$ = number of categories of input signal;
   $K$ = number of antigen in tissue antigen vector;
   $L$ = number of DC cycles;
   $M$ = number of DCs in population;
   $N$ = DC antigen vector size ;
   $P$ = number of output signals per DC;
   $Q$ = number of antigens sampled per DC, per cycle;
   $R$ = number of DC antigen receptors;
   $T_{max}$ = tissue antigen vector size.

– Data Structures:
   $T = \{S, A\}$ - the tissue;
   $S$ = tissue signal matrix;
   $s_{ij}$ = a signal index i, category j in the signal matrix S;
   $A$ = tissue antigen vector;
   $a_k$ = antigen k in the tissue antigen vector;
   $DC_m = \{s(m), a(m), \overline{o_p}(m), t_m\}$ - a DC within the population;
   $s(m)$ = signal matrix of $DC_m$;
   $a(m)$ = antigen vector of $DC_m$;
   $o_p(m)$ = output signal p of $DC_m$;
   $\overline{o_p}(m)$ = cumulative output signal p of $DC_m$;
   $t_m$ = migration threshold of $DC_m$;
   $w_{ijp}$ = transforming weight from $s_{ij}$ to $o_p$.

$$o_p(m) = \sum_i \sum_{j \neq 3} w_{ijp} s_{ij}(m) \quad \forall p \tag{2}$$

Each $DC_m$ transforms each value of s(m) to $o_p$(m). In Equation 2, a specific example is given for use with four input signals, with one signal per category and consists of some additional components. Additionally, the j = 3 component implies that signal category index is not summed with the other three signal categories i.e. inflammation is not treated in the same manner as the other signals, as shown in this equation. The interrelationships between the weights, determined through practical immunology, are shown in Table 2.

The tissue has containers for signal and antigen values, namely S and A. In this example version of the DCA, there are four categories of signal (J = 3) and one signal per category (I = 0). In this instantiation $s_{0,0}$ = PAMP signals, $s_{0,1}$ = danger signals and $s_{0,2}$ = safe signals and $s_{0,3}$ = inflammation.

Table 2  Derivation and interrelationship between weights in the signal processing equation, where the values of the PAMP weights are used to create all other weights relative to the PAMP weight. $W1$ is the weight to transform the PAMP signal to the CSM output signal and $W2$ is the weight to transform the PAMP signal to the mature output signal.

| $w_{ijp}$ | $j=0$ | $j=1$ | $j=2$ |
|---|---|---|---|
| $p=0$ | $W1$ | $\frac{W1}{2}$ | $W1*1.5$ |
| $p=1$ | $0$ | $0$ | $1$ |
| $p=2$ | $W2$ | $\frac{W2}{2}$ | $-W2*1.5$ |

An antigen store is constructed for use within the tissue cycle where all DCs in the population can collect antigen.

The cell update component maintains all DC data structures including the DC population of set size $M$. Each DC has an input signal matrix, antigen vector, cumulative output signals and migration threshold. The internal values of $DC_m$ are updated, based on current data in the tissue signal matrix and antigen vector. The DC input signals, $s(m)$ use the identical mapping for signal categories as tissue $s$ and are updated every cell cycle iteration. Each $s(m)$ for $DC_m$ is updated via an overwrite every cell cycle. These values are used to calculate output signal values, $o_{p(m)}$, for $DC_m$, which are added cumulatively over a number of cell cycles to form $\overline{o}_p(m)$, where $p=0$ is costimulatory value, $p=1$ is the semi-mature DC output signal and $p=2$ is the mature DC output signal. With each cell update, DCs sample $R$ antigens from the tissue antigen vector $A$.

After the internal values of a DC are updated, $o_0$ is assessed against $t_m$ the cell's migration threshold. If $o_0$ is greater than $t_m$, the DC is 'removed' from the tissue. Here, 'remove' means that the DC is de-allocated the receptors needed to sample the signal matrix and to collect antigen. On the next update cycle, the remaining output signals are checked and the analysis procedure is initiated.

In this implementation, each DC is assigned a random value for $t_m$, within a specified range. The random value adds some diversity to the DC population. The value of $o_0$ is increased on exposure to any signal and proportionately to the strength of the input signal. By using randomly assigned migration thresholds, each DC samples the signal matrix a different number of times throughout its lifetime. Some exist for a short period sampling once or twice, others can persist for longer, dependent on the strength of the signals. This creates a variable time window effect for the sampling.

Pseudocode for this specific instantiation of the DCA is given in Algorithm 1. This pseudocode shows both the update of the tissue and the individual DCs. The stages of the algorithm are shown, namely initialisation, update and analysis. While this provides the detail of the DC update mechanisms, this pseudocode does not encapsulate the asynchronous nature of the update stages. As `libtissue` is a multithreaded framework, the three updates are controlled by three different processes, therefore, the three updates occur asynchronously. This architecture is particularly suited for real time data processing as updates occur as and when they are required.

## 3.6 Antigen Aggregation

Once $DC_m$ has been removed from the population, the contents of $a(m)$ and values $\overline{o}_p(m)$ are logged to a file for the aggregation stage. Once completed, $s(m)$, $a(m)$ and $\overline{o}_p(m)$ are all reset and $DC_m$ is returned to the sampling population. The re-cycling of DCs continues until the stopping condition is met ($l=L$). Once all data has been processed by the DCs, the output log of antigen-plus-context is analysed.

The same antigen is presented multiple times with different context values. This information is recorded in a log file. The total fraction of mature DCs presenting said antigen (where $\overline{o}_1 > \overline{o}_2$) is divided by the total amount of times the antigen was presented namely $\overline{o}_1/(\overline{o}_1+\overline{o}_2)$. This is used to calculate the mean MCAV. Equation 1 describes this process.

```
input  : T = {S, A}
output: a and context
create DCs;
initialise parameters {I, J, K, L, M, N, O, P, Q};
for l < L do
    update A and S;
    for m = 0 to M do
        for n = 0 to Q do
        |   DC_m samples Q antigen from A;
        end
        for i = 0 to I and j = 0 to J do
        |   s_ij^DC = s_ij;
        end
        for n = 0 to N do
        |   DC_m processes a_n(m);
        end
        for p to P do
        |   compute o_p;
        |   o_p(m) = o_p(m) + o_p;
        end
        if o_0(m) > t_m then
        |   DC_m removed from population;
        |   DC_m migrate, print antigen and context;
        |   DC_m reset antigen vector and all signals;
        end
    end
    l++;
end
analyse antigen and calculate MCAV
```

Algorithm 1: Pseudocode of the implemented DCA

## 3.7 Signals and Antigen

An integral part of DC function is the ability to combine multiple signals to produce context information. The semantics of the different categories of signal are derived from the study of the influence of the different signals on DCs *in vitro*. Definitions of the characteristics of each signal category are given below, with an example of an actual signal per category. This categorisation forms the signal selection schema. Any number of sources of information can be mapped using the outlined principles.

– PAMP - $s_{i0}$, *e.g. the number of error messages generated per second by a failed network connection:*
  1. a signature of abnormal behaviour, e.g. an error message;
  2. a high degree of confidence of abnormality associated with an increase in this signal strength.
– Danger signal - $s_{i1}$, *e.g. the number of transmitted network packets per second:*
  1. measure of an attribute which significantly increases in response to abnormal behaviour;
  2. a moderate degree of confidence of abnormality with increased level of this signal, though a low signal strength can represent normal behaviour.
– Safe signal - $s_{i2}$ *e.g. the inverse rate of change of number of network packets per second. A high rate of change equals a low safe signal level and vice versa:*
  1. a confident indicator of normal behaviour in a predictable manner or a measure of steady-behaviour;
  2. measure of an attribute which increases signal concentration due to the lack of change in strength.

Signals, though interesting, are inconsequential without antigen. To a DC, antigen is an element which is carried and presented to a T-cell, without regard for the structure of the antigen. Antigen *is* the data to be classified and works well in the form of an identifier, be it an anomalous process ID or the ID of a data item [26]. At this stage, minimal antigen processing is performed and the antigen presented is an identical copy of the antigen collected. Detection is performed through the correlation of antigen with fused signals.

# 4 Self-Organizing Maps

## 4.1 Biological Inspiration for SOM

Various properties of the brain were used as an inspiration for a large set of algorithms and computational theories known as neural networks. Such algorithms have shown to be successful, however a vital aspect of biological neural networks was omitted in the algorithm's development. This was the notion of self-organization and spatial organization of information within the brain. In 1981 Kohonen proposed a method which takes into account these two biological properties and presented them in his SOM algorithm [46].

The SOM algorithm generates, usually, two dimensional maps representing a scaled version of n-dimensional data used as the input to the algorithm. These maps can be thought of as "neural networks" in the same sense as SOM's traditional rivals, artificial neural networks (ANNs). This is due to the algorithm's inspiration from the way that mammalian brains are structured and operate in a data reducing and self-organised fashion. Traditional ANNs originated from the functionality and interoperability of neurons within the brain. The SOM algorithm on the other hand was inspired by the existence of many kinds of "maps" within the brain that represent spatially organised responses. An example from the biological domain is the somatotopic map within the human brain, containing a representation of the body and its adjacent and topographically almost identical motor map responsible for the mediation of muscle activity [47].

This spatial arrangement is vital for the correct functioning of the central nervous system [40]. This is because similar types of information (usually sensory information) are held in close spatial proximity to each other in order for successful information fusion to take place as well as to minimise the distance when neurons with similar tasks communicate. For example sensory information of the leg lies next to sensory information of the sole.

The fact that similarities in the input signals are converted into spatial relationships among the responding neurons provides the brain with an abstraction ability that suppresses trivial detail and only maps most important properties and features along the dimensions of the brain's map [61].

## 4.2 SOM Algorithm Overview

As the algorithm represents the above described functionality, it contains numerous methods that achieve properties similar to the biological system. The SOM algorithm comprises of competitive learning, self-organization, multidimensional scaling, global and local ordering of the generated map and its adaptation.

There are two high-level stages of the algorithm that ensure a successful creation of a map. The first stage is the *global ordering* stage in which we start with a map of predefined size with neurons of random nature and using competitive learning and a method of self-organization, the algorithm produces a rough estimation of the topography of the map based on the input data. Once a desired number of input data is used for such estimation, the algorithm proceeds to the *fine-tuning* stage, where the effect of the input data on the topography of the map is monotonically decreasing with time, while individual neurons and their close topological neighbours are sensitised and thus fine tuned to the present input.

The original algorithm developed by Kohonen comprises of initialisation followed by three vital steps which are repeated until a condition is met:

- Choice of stimulus
- Response
- Adaptation

Each of these steps are described in detail in the next section.

## 4.3 Algorithmic Detail and Implementation

A number of existing software packages that contain an implementation of the SOM algorithm are available, however these are not ideal for some types of security research, such as real-time detection, due to their limited speed and integratability. For this reason a C++ based implementation is developed according to the original incremental SOM algorithm as described by Kohonen [48]. In this section a detailed analysis of the implemented algorithm and its step by step functional description follows.

### 4.3.1 Initialisation

A number of parameters have to be chosen before the algorithm is to begin execution. These include the size of the map, its shape, the distance measure used for comparing how similar nodes are, to each other and to the input feature vectors, as well as the kernel function used for the training of the map. Kohonen suggested recommended values for these parameters [47], which are used throughout our experiments. The values used in this paper are described in Section 5, found in Table 8. Once these parameters are chosen, a map is created of the predefined size, populated with nodes, each of which is assigned a vector of random values, $w_i$, where i denotes node to which vector w belongs.

### 4.3.2 Stimulus Selection

The next step in the SOM algorithm is the selection of the stimulus that is to be used for the generation of the map. This is done by randomly selecting a subset of input feature vectors from a training data set and presenting each input feature vector, x, to the map, one item per epoch. An epoch represents one complete computation of the three vital steps of the algorithm.

### 4.3.3 Response

At this stage the algorithm takes the presented input, x and compares it against every node i within the map by means of a distance measure between x and each nodes' weight vector $w_i$. For example this can be the Euclidean distance measure shown in Equation 3, where $||.||$ is the Euclidean norm and $w_i$ is the weight vector of node i. This way a winning node can be determined by finding a node within the map with the smallest Euclidean distance from the presented vector x, here signified by c.

$$c = \operatorname{argmin}\{||x - w_i||\} \tag{3}$$

### 4.3.4 Adaptation

Adaptation is the step where the winning node is adjusted to be slightly more similar to the input x. This is achieved by using a kernel function, such as the Gaussian function ($h_{ci}$) as seen in Equation 4 as part of a learning process.

$$h_{ci}(t) = \alpha(t).\exp\left(-\frac{||r_c - r_i||^2}{2\sigma^2(t)}\right) \tag{4}$$

In the above function, $\alpha(t)$ denotes a "learning-rate factor" and $\sigma(t)$ denotes the width of the neighbourhood affected by the Gaussian function. Both of these parameters decrease monotonically over time (t). During the first 1,000 steps, $\alpha(t)$ should have reasonably high values (e.g. close to 1). This is called the *global ordering* stage and is responsible for proper ordering of $w_i$. For the remaining steps, $\alpha(t)$ should attain reasonably small values ($\geq 0.2$), as this is the *fine-tuning* stage where only fine adjustments to the map are performed. Both $r_c$ and $r_i$ are location vectors of the winner node (denoted by subscript c) and i respectively, containing information about a node's location within the map.

$$w_i(t + 1) = w_i(t) + h_{ci}(t)[x(t) - w_i(t)] \tag{5}$$

The learning function itself is shown in Equation 5. Here the Gaussian kernel function $h_{ci}$ is responsible for the adjustment of all nodes according to the input feature vector x and each node's distance from the winning node. This whole adaptation step is the vital part of the SOM algorithm that is responsible for the algorithm's self-organisational properties.

### 4.3.5 Repetition

Stimulus selection, Response and Adaptation are repeated a desired number of times or until a map of sufficient quality is generated. For our experiment this was set to a value suggested by Kohonen [48]. He states that the number of steps should be at least 500 times the number of map units. For this reason 100,000 epochs were used in our experiments. Another possible mechanism for the termination of the algorithm is the calculation of the quantization error, which is the mean of $\|x - w_c\|$ over the training data. Once the overall quantization error falls below a certain threshold, the execution of the algorithm can stop as an acceptable lower dimensional representation of the input data has been generated.

## 5 Experimental Comparison

### 5.1 Scenarios

For the experiments in this paper, two data sets are compiled, collected using a system of signal collection scripts with raw input signal data collected from the Linux `/proc` filesystem. One data set is termed passive normal (PN) and contains a SYN scan performed without normal processes invoked by a user i.e. scan and shell processes. The second data set is termed active normal (AN). This data set contains an identical SYN scan but is combined with simultaneous instances of normal programs which are used actively by a user throughout the duration of the session i.e. scan, shell processes and a firefox web browser.

Block scans are conducted across 254 IP addresses connecting to multiple ports on each host successfully probed. Approximately 70 hosts out of the 254 addresses scanned are available at any one instance during the scan. For these scenarios, the DCA resides on a client machine which is connected to the main network. As the scanned local hosts are part of a university network and the availability of the hosts is beyond direct control, with the exception of the host on which the DCA monitors. The scan performed in both data sets is a standard SYN scan, with a fast probe sending rate ( <0.1 seconds per probe), facilitated through the use of the popular scanning tool, nmap [17]. The command invoked to perform the SYN scan using nmap is "`nmap -sS -v xxx.xxx.xxx.1-254`".

The AN data set is 7,000 seconds in duration, with 'normal' antigen generated by running a web browser over a separate remote ssh session. During browsing, multiple downloads, chat sessions and the receipt of e-mail occur representing different patterns of network behaviour. This actively generated network traffic is provided to observe if the algorithms can differentiate between two highly active processes which run simultaneously and modify the networking behaviour of the victim host. Having both normal and anomalous processes running via the monitored ssh demon may make the detection of the scan more difficult. This may increase the MCAV for the normal processes as the DCA relies on the temporal correlation of signals and antigen to perform detection.

The PN data set is also 7,000 seconds in duration and comprises of a SYN scan and its pseudo-terminal slave (pts) parent process as anomalous examples. The ssh demon process acts as normal antigen and is needed to facilitate the remote login. In addition, a firefox browser runs throughout the session, but the system calls are run locally and not through the ssh demon. Therefore this does not form antigen, but can influence the input signal data.

### 5.2 Data Pre-processing and Signals

The DCA relies on correct mapping of signals, ensured through the examination of preliminary samples of input data. For the detection of SYN scans, seven signals are derived from behavioural attributes of the victim host: two PAMPs, two danger signals, two safe signals and one inflammatory signal. Having multiple signals per category may make the DCA more robust against random network fluctuations or conversely could impede classification through conflicting inputs.

The PAMP signals (PAMP-1 and PAMP-2) are both taken from data sources which indicate a scan specifically. Danger signals (DS-1 and DS-2) are derived from attributes which represent changes in behaviour. Safe signals (SS-1 and SS-2) are also derived from changes in behaviour, but high safe signal values are shown when the changes are small in magnitude. The inflammatory signal is simplified to a binary signal i.e. inflammation present (1) or not (0). All PAMPs, danger and safe signals are normalised within a range of zero to 100. A sketch of the input signals throughout the duration of the two sessions are shown in Figures 7, 8 and 9 for the AN data set and in Figure 10, 11 and 12 for the PN data set.

To devise a set of appropriate signals a number of preliminary experiments must be performed, in addition to the acquisition of knowledge regarding the effects of scanning and normal networking usage within a host. Initially a plethora of system variables are monitored under a variety of situations. The signals used in this experiment are network based attributes. This kind of system data appears to be the most variable under scanning conditions. Once the candidate signals are selected, they are then categorised using the general principles of signal selection i.e. PAMPs are signature-based, danger signals are normal at low values and anomalous at high values etc. Following the categorisation, the raw values of signals are transformed into normalised signals.

*PAMP-1* is the number of ICMP 'destination unreachable' (DU) error messages received per second. Scanning IP addresses attached to hosts which are firewalled against ICMP packets generate these error messages in response to probing. This signal is shown to be useful in detecting ping scans and may also be important for the detection of SYN scans, as an initial ping scan is performed to find running hosts. In this experiment, the number of ICMP messages generated is significantly less than observed with a ping scan. To account for this, normalisation of this signal includes multiplying the raw signal value by five, capped at a value of 100 (equivalent to 20 DU errors per second). This process is represented in Equation 6 where raw is the unmodified system data and signal represents the normalised output signal. These terms apply to all equations described within this section.

$$signal = \min\{100, raw * 5\} \qquad (6)$$

*PAMP-2* is the number of TCP reset packets sent and received per second. Due to the nature of the scan, a volume of RST packets are created in both port status cases; they are generated from the scanning host if ports are open and are generated by the remote hosts if ports are closed. RST packets are not usually present in any considerable volume, so their increased frequency is a likely sign of scanning activity. This signal is normalised linearly, with a maximum cap set at 100 RSTs per second. This normalisation process is shown in Equation 7.

$$signal = \min\{100, raw\} \qquad (7)$$

*DS-1*, the first danger signal is derived from the number of network packets sent per second. Previous experiments with this signal data indicate that it is useful for the detection of outbound scans [27]. A different approach is taken for the normalisation of this signal. A sigmoid function is used to emphasise the differences in observed rate, making the range of 100 to 700 packets per second more sensitive. This sensitive range is determined through preliminary data analysis of host behaviour during scans and normal use, with 750 packets per second found to be the median value across the plethora of preliminary data. This function makes the system less sensitive to fluctuations under 100 packets per second, whilst keeping the sensitivity of the higher values. A cap is set at 1500 packets per second, resulting in a signal range between 0 and 100. The general sigmoid function used for this transformation is shown in Equation 8, with the specific function shown in Equation 9.

$$f(x) = \frac{1}{1 + e^{-x}} \qquad (8)$$

$$signal = min\{ \left( \frac{1}{1 + 2^{(7.5 - \frac{raw}{100})}} \right) * 100, 100 \} \qquad (9)$$

*DS-2* is derived from the ratio of TCP packets to all other packets processed by the network card of the scanning host. This may prove useful as during SYN scans there is a burst of traffic comprised of almost entirely TCP type packets, which is not usually observed under normal conditions. The ratio is normalised through multiplication by 100, to give this signal the same range as DS-1. This procedure is shown in Equation 10.

$$signal = \left( \frac{rawSignalTcp}{rawSignalAllPkts} \right) * 100 \qquad (10)$$

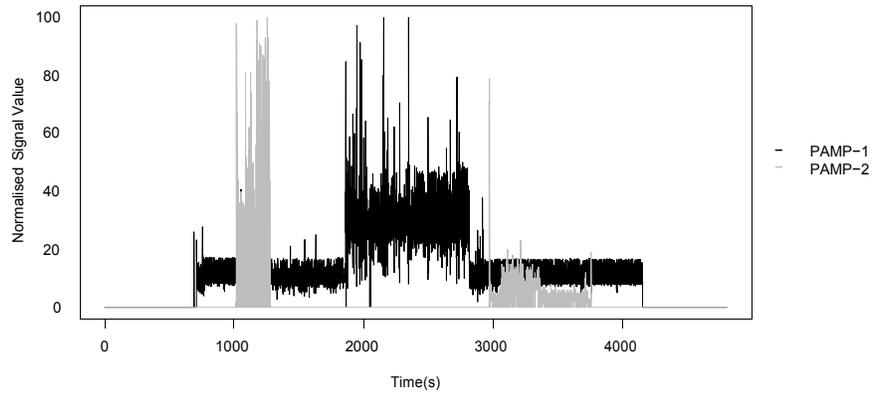

Fig. 7 Line graph of the PAMP (PAMP-1, PAMP-2) signals which constitute the AN (active normal) data set.

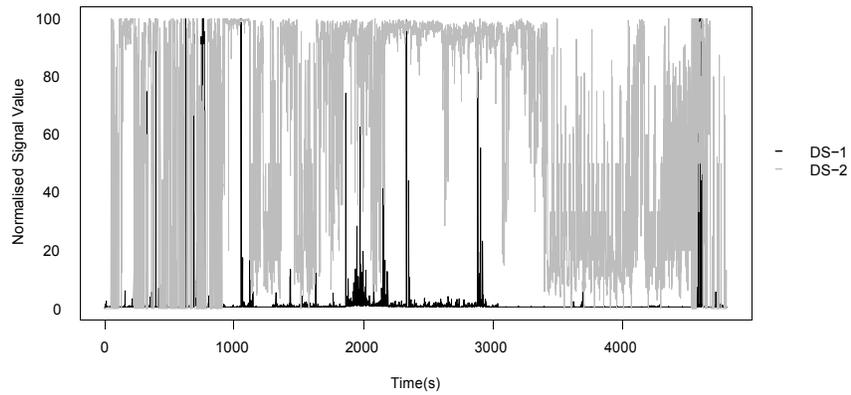

Fig. 8 Line graph of the danger (DS-1, DS-2) signals which constitute the AN (active normal) data set.

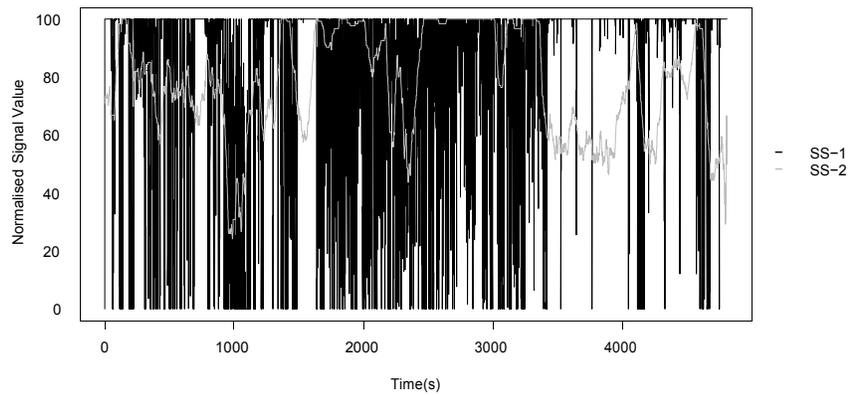

Fig. 9 Line graph of safe (SS-1, SS-2) signals which constitute the AN (active normal) data set.

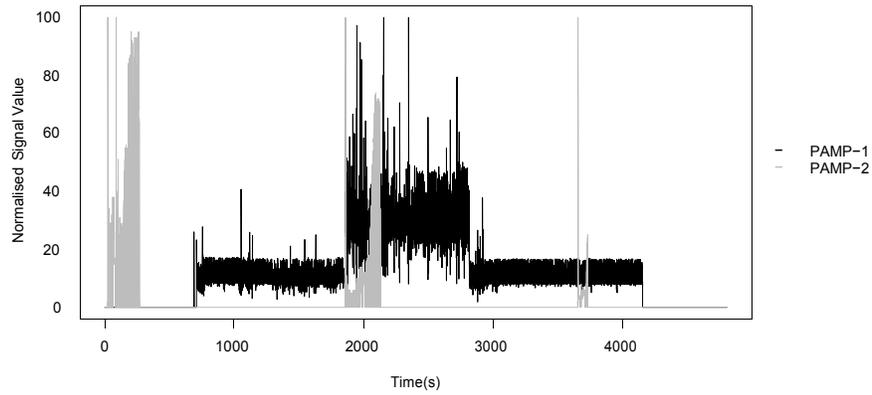

Fig. 10 Line graph of the PAMP (PAMP-1, PAMP-2) signals which constitute the PN (passive normal) data set.

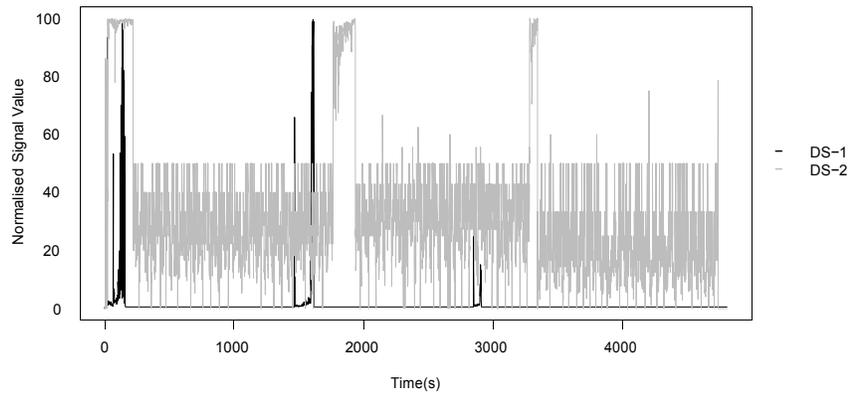

Fig. 11 Line graph of the danger (DS-1, DS-2) signals which constitute the PN (passive normal) data set.

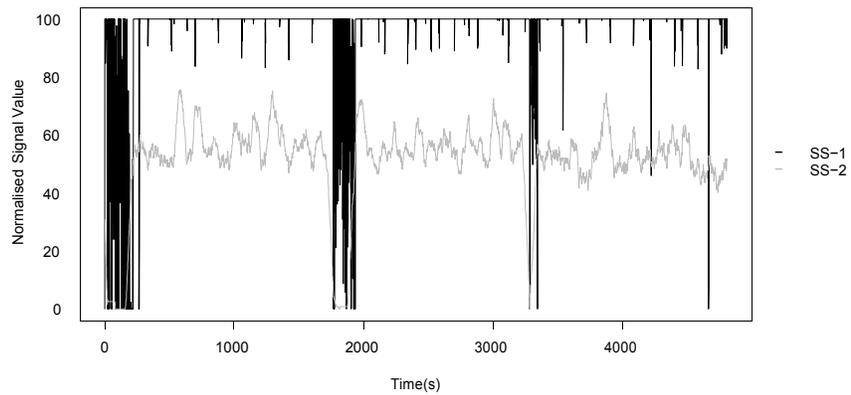

Fig. 12 Line graph of the safe (SS-1, SS-2) signals which constitute the PN (passive normal) data set.

**Table 3** Ranges used in the normalisation function of SS-2.

| Range | Signal Value |
|-------|--------------|
| 40 - 45 | 0 |
| 46- 50 | 10 |
| 51- 60 | 50 |
| 61 + | 100 |

*SS-1* is the rate of change of network packet sending per second. Safe signals are implemented to counteract the effects of the other signals, hopefully reducing the number of false positive antigen types. High values of this signal are achieved if the rate is low and vice versa. This implies that a large volume of packets can be legitimate, as long as the rate at which the packets are sent remains constant. The value for the rate of change can be calculated from the raw DS-1 signal value, though conveniently, the `proc` file system also generates a moving-average representation of the rate of change of packets per second (over 2 seconds). This raw value can be normalised between values of 10 and 100 and inverted so that the safe signal value decreases as the raw signal value increases. This normalisation process is described in Equation 11.

$$\text{signal} = \min\{100, \max\{0, (100 - \text{raw}) * \frac{10}{9}\}\} \tag{11}$$

*SS-2* is based on the observation that during SYN scans the average network packet size reduces to a size of 40 bytes, with a low standard deviation. Preliminary observations under normal conditions show that the average packet size for normal traffic is within a range of 70 and 90 bytes. A step function is implemented to derive this signal with transformation values presented in Table 3. Preliminary experiments have also shown that a moving average is needed to increase the sensitivity of this signal. This average is created over a 60 second period.

The inflammatory signal is binary and based on the presence of remote root logins. If a root log-in is detected through the monitored ssh demon, this signal is assigned a value of one. When used in the signal processing equation, this multiplies the resultant values of the other signals by two, acting as an amplifier for all other signals including the safe signal. This signal may be useful as to perform a SYN scan the invocation has to come from a user with root privileges. While this is a very important feature of the SYN scan process, it is not suitable for use as a PAMP signal as it can be easily spoofed. Thorough analysis of the relationship between inflammation and the DCA is outside of the scope of this paper and may feature in future work. While this signal influences the rates of migration of the DCs, it does not influence the rates of detection as the addition of this signal increases the output signal values for both the semi-mature output ($o_1$) and the mature output ($o_2$) signals.

As shown in Figure 7-9 and 10-12, the AN signals are more variable than the PN signals, as many more processes run during the AN session. In the AN session, the nmap scan is invoked at 651 seconds. Signals PAMP-1, PAMP-2 and DS-2 clearly change for the duration of the scan. The remaining signals are less clear, though some evidence of changes throughout the scan duration is shown. The changes are transient and localised in particular to the beginning of the scan, when the majority of probes are sent to other hosts.

The signals of the PN data set are less noisy. Analysis of input antigen confirms nearly 99% of these antigen belong to the anomalous pts and nmap processes. PAMP-1 and PAMP-2 are responsive to the scan, as shown by their rapid decline towards the end of the scanning period, at 5,500 seconds. Changes in DS-1 are more pronounced in the PN data set, yet the magnitude of this signal is smaller than expected. DS-2 appears to be highly correlated with the scan, yielding values of over 20 throughout the scan duration. SS-1 performs poorly and only decreases in response to the scan in a few select places. SS-2 falls sharply in the middle of the scan, as predicted, but otherwise remains at a constant level of 60 even after the scan has finished.

## 5.3 Antigen

Process identification numbers (PIDs) form the antigen and are generated each time a system call is invoked. To provide antigen, all remote sessions facilitated by ssh are monitored for this experiment.

**Table 4** Summary statistics of the frequency of system calls for the nmap and firefox processes.

| Statistic | Nmap | Firefox |
|---|---|---|
| Mean | 2445 | 880 |
| Standard Deviation | 1243 | 840 |
| 1st Quartile | 1796 | 76 |
| 2nd Quartile (Median) | 2106 | 792 |
| 3rd Quartile | 2894 | 1479 |
| 4th Quartile (Total) | 11758 | 7156 |
| Mean/SD | 1.97 | 1.05 |

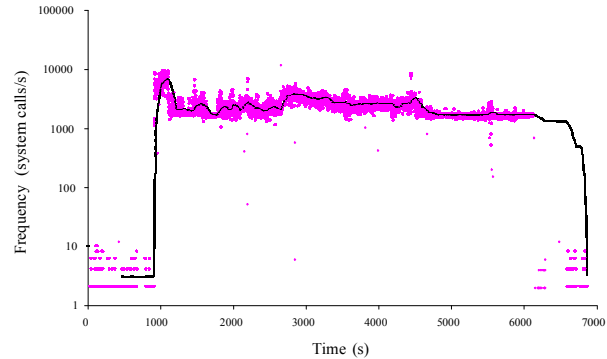

**Fig. 13** The frequency of system calls invoked by the nmap scan process for the AN data set. The trendline represents a moving average over 100 data points. As this figure is presented on a logarithmic scale, the absence of points indicates a frequency of zero system calls for that particular second.

Using multiple system calls with identical PIDs allows for the aggregate antigen sampling method, having multiple antigen per antigen type. This allows for the detection of active processes when changes in signal values are observed. This technique is a form of process anomaly detection, but the actual structure of the PID is not important in terms of its classification, i.e. no pattern matching is performed, on the actual value of the PIDs: it is a label for the purpose of process identification.

A graph of the frequency of system calls invoked per second for the AN data set by the nmap process is shown in Figure 13 and for the firefox process in Figure 14. In these two figures, individual points represent the frequency of system calls per second, while the trendline represents a moving average over 100 points. Summary statistics of the system call data are given in Table 4, which are generated across the entire session for both processes.

The mean/median frequency of system calls for the nmap process is higher than the firefox process. To assess which process is more variable, the means are divided by the standard deviations, as shown in the summary table. This value is larger in the case of nmap than for firefox. This indicates that relatively, the standard deviation of the firefox process is larger in comparison to the mean than that of the nmap process. The various proportions of input system calls are represented as a chart in Figure 15 and shows that the nmap process invokes the majority of system calls in the AN data set.

## 5.4 Special Cases for SOM

For use with the SOM both antigen and signal data must be correlated explicitly - this is a feature of the DCA that cannot be translated for use with the SOM. To achieve this coupling, the timestamps of the signal and antigen data are used. Each antigen is assigned the set of signals which occurs within one second of the antigen generation. This results in what resembles a standard feature vector, where antigens are the data item ID and signals form seven attributes. In addition to this pairing, the SOM requires training data, to assign nodes and to create the map. Unlike the DCA which uses expert knowledge, SOM requires training data sets and therefore ten sets are constructed. Each training set consists of a diverse number of normal processes which use the networking facility of the host machine.

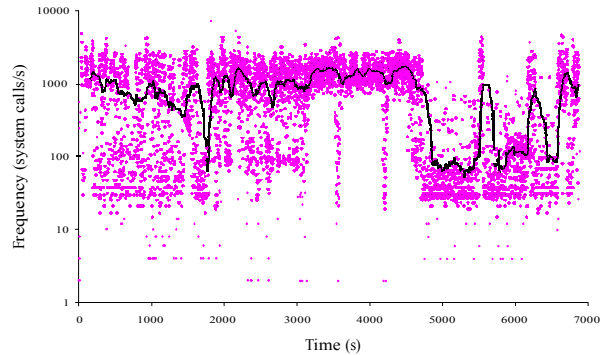

Fig. 14 The frequency of system calls invoked by the firefox web browser process for the AN data set. The trendline represents a moving average over 100 data points. This figure is presented on a logarithmic scale and the absence of points indicates a frequency of zero system calls for that particular second.

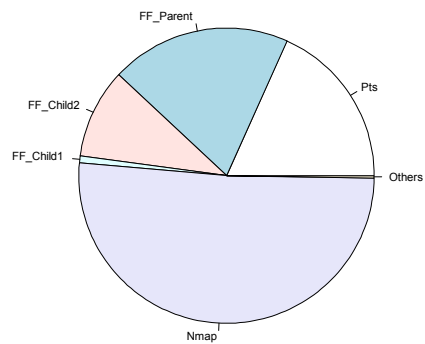

Fig. 15 Proportionate chart of antigen per processes as input data for the active normal data set, where FF_Parent is the parent firefox process and FF_Child1 and FF_Child2 are the forked children of the parent process.

### 5.4.1 Training

Due to the SOM algorithm's unsupervised learning nature, training data is required in order to generate a map that is representative of normal behaviour. As described earlier this is not necessary for the DCA as there is no training stage. A set of ten sessions of normal activity are generated in an identical manner to the testing data except for the fact that no antigens are generated as the SOM is only trained on signals. Antigen themselves bring no benefit to the SOM in the training stage. Each session contains approximately 60 minutes worth of data. The data contains normal activity such as internet browsing, chatting, file transfer and other activities performed by a standard user.

Once the data is collected, it is combined into one data set, which is subsequently used as the input into the SOM algorithm. Input feature vectors are then selected from this data set at random and presented to the map for computation. The training results in a map which can be seen in Figure 16. In this example, the brighter the colour, the greater the dissimilarity of neighbouring nodes, with the map representing four clusters. This shows one of the maps that was generated throughout our experiments. Ten runs were performed, both for training and detection.

### 5.4.2 Detection

The SOM algorithm itself cannot perform anomaly detection without any further processing. A meaningful way has to be devised in order to be able to classify and make a decision whether a data item or a set of items are anomalous. The aim of our experiments was to detect whether a whole process is anomalous or not. Thus a method for process, rather than signal set anomaly detection calculation had to be developed.

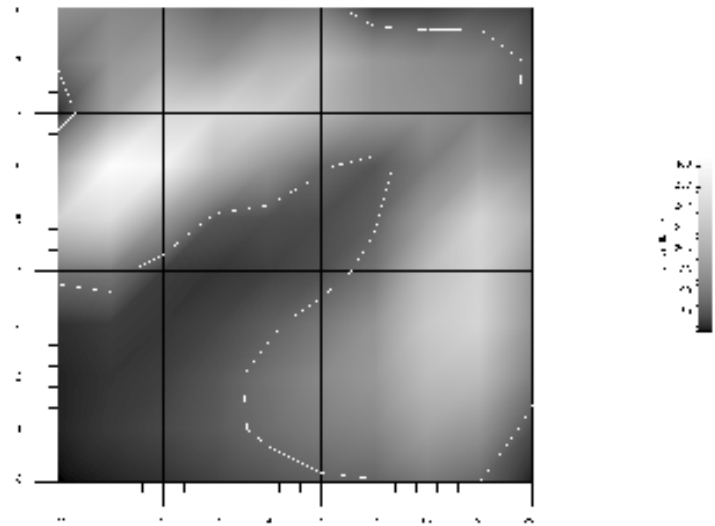

Fig. 16 An example map generated by the original SOM algorithm.

For informational purposes a simple anomaly detection can be performed on our data by classifying sets of signals only. This can be done by firstly using the calculation in Equation 3, which will determine the winning node within our trained map. We will call this the *Best Matching Unit* (BMU). Once the BMU is found, the actual Euclidean distance between the currently observed vector of signals and the BMU is calculated. The most trivial anomaly detection is done by choosing a threshold for this dissimilarity. If the currently observed item is too different from the BMU, then it is deemed anomalous.

In order to perform process anomaly detection, antigen information has to be correlated with signals from the testing data. As the SOM is trained on signals only, antigens (PIDs) need to be correctly correlated with the right signals in order to be able to link anomalous sets of signals to their respective initiators (processes). Initial correlation is done by synchronizing antigens with signals using timestamp information. Any antigen with timestamp t is assigned a signal set at time t, and for the purpose of synchronisation t is measured in seconds. Once this synchronisation takes place, the signal set - antigen coupling is assessed for its anomaly level using the BMU technique described previously.

As explained later in this section, the output antigen from the DCA are 'segmented' into fixed sized sets for analysis. These 'antigen segments' consist of a specified number of output antigen used to form the MCAV values. Multiple MCAV values are produced as a result of this procedure, which may add additional sensitivity. To get a more meaningful and easily comparable results to the DCA, a segmentation post-processing needs to be employed. This is achieved by selecting a segment size z which generates the same number of segments as the DCA. The reason behind the differing segment sizes is the way that synchronisation of signal and antigen is achieved in the SOM experiments compared to the way it is achieved by the DCA. The DCA correlates antigen and signal as part of the algorithm's operation whereas the signal and antigen need to be manually correlated at a pre-processing stage before analysis using the SOM can occur. In our case z = 1,800; 18,000; 180,000 and 1,800,000. These segment sizes produce the same number of segments as segment sizes of 100; 1,000; 10,000 and 100,000 respectively, used by the DCA.

Further post-processing is done by using a binary discrimination of how anomalous a signal set - antigen coupling is. It was observed that the median dissimilarity value between an input feature vector and the BMU is approximately 65. Thus this value is chosen as the threshold for deciding whether a signal set-antigen coupling should be deemed anomalous and thus assigned a value of 1 or deemed normal and assigned a value of 0. These binary anomaly values are then used for the calculation of the mean process anomaly value.

Table 5 Default parameter settings for the DCA, chosen following the sensitivity analysis performed previously [24]. Values shown indexed from zero.

| Name | Symbol | Value |
|---|---|---|
| Number of signals per category | I | 0 |
| Number of signals categories | J | 3 |
| Max number of tissue antigen | K | 499 |
| Number of cells | M | 99 |
| Max number of antigen per DC | N | 49 |
| Number of output signals per DC | P | 3 |
| Number of DC antigen receptors | Q | 9 |

Segment generation is achieved by calculating the anomaly value for each signal set - antigen coupling as described above, followed by further analysis of these results per the various antigen segment sizes of z. Thus for $z = 1,800$, the first 1,800 data items were processed in the following way. For each unique antigen a mean anomaly value is calculated for that given segment. Once this calculation is performed, each segment contains a list of unique antigens with their associated mean anomaly value. We call this the *Mean Best Matching Unit* (MBMU) value. This value gives an indication of how anomalous a process is in the given antigen segment.

## 5.5 Experiments

Two null hypotheses are used in these SYN scan experiments:

1. H1: modification of the antigen segment size will not influence the resultant anomaly coefficient values for both normal and anomalous processes
2. H2: the DCA and the SOM will produce results which are not statistically different

Both the AN and PN data sets are used in this series. Each data set is collected using a real-time version of the DCA, which also provides the opportunity to verify the detection rate results before an in-depth analysis is performed. Ten runs are performed for each data set and the mean MCAV values / MBMU values across the ten runs are recorded. Unlike previous experiments with the DCA, no system parameters are varied for these experiments. Instead the number of antigen used in the MCAV calculation is varied.

For previous DCA application of ping scan detection, the sessions used are at most one minute in duration and generate approximately 3,000 antigen. However, the AN and PN data sets generate in excess of 100,000 antigen per process therefore analysis once all data is processed may not provide meaningful results. Generating the output coefficient value at the end of the 7,000 second session is too insensitive to changes which happen over a matter of seconds or minutes. Figures are generated showing the varying coefficients per process as the session proceeds and comparisons drawn between the performance for the two data sets.

In terms of assessment, the PIDs with the highest volume of antigen output are used as the processes of interest. For the passive normal data set these processes are the nmap scan process and the pts parent process of the SYN scan. The processes of interest for the active normal data set include the nmap scan, pts process and the firefox browser. Graphs are generated showing the MCAV and MBMU for each process of interest per z antigen presented, for the duration of the experiments for example z=100. Higher values of MCAV are expected for the SYN scan process and its parent process the ssh demon, than for the firefox browser. It is expected that smaller values of z will yield an improvement in the precision and accuracy of the detection, though when $z = 100$, the system may be too sensitive and an element of tolerance to false positives could be lost. The variants of z for the DCA are presented in Table 6.

All data sets are collected and analyses with the DCA are performed on an AMD Athlon 1 GHz Debian linux machine (kernel 2.4.10). A 2.66 GHz Intel Core 2 Duo Ubuntu linux machine (kernel 2.6.22) is used for the SOM processing. Data generated from the DCA machine is used in SOM processing to avoid kernel discrepancies. The DCA is implemented in C (gcc 4.0.2) with the SOM implemented in C++ (g++ 4.1.3). All raw signals are derived using signal collection scripts, with values taken from the /proc filesystem (PAMP-1, DS-1, SS-1, I), the tcpstat linux utility (DS-2, SS-2) and a custom developed packet sniffer (PAMP-2). The system parameters used with the DCA are shown in Table 5, derived as a result of previous DCA sensitivity analysis [24]. Weights for the signal processing of these data are shown in Table 7. These weights provide a shorter time-window for the duration of signal sampling per DC, shown to be advantageous for the detection of ping scans. The migration threshold $t_m$ is assigned a value of 60 +/− 30 which is the median signal value observed across all signals. The parameters used for the SOM are shown in Table 8 and are chosen based on recommended values as proposed by Kohonen [47].

Table 6  The various sizes of antigen segment z used for the DCA and SOM respectively.

| DCA | SOM |
|---|---|
| 100 | 1,800 |
| 1,000 | 18,000 |
| 10,000 | 180,000 |
| 100,000 | 1,800,000 |
| 1,000,000 | - |

Table 7  Weights used for signal processing, where $j$ represents the input signal category, i represents an instance of a signal within signal category $j$ and p is the corresponding output signal.

| $w_{ijp}$ | $j = 0$ | $j = 1$ | $j = 2$ |
|---|---|---|---|
| $p = 0$ | 4 | 2 | 6 |
| $p = 1$ | 0 | 0 | 1 |
| $p = 2$ | 8 | 4 | -12 |

Table 8  SOM parameters, where $\alpha$ is the "learning-rate factor" and $t$ is the current epoch.

| Parameter | Global Ordering | Fine-Tuning |
|---|---|---|
| SOM size | 10*10 | - |
| Epoch Limit | 100,000 | - |
| Initial $\alpha$ | 0.9 | 0.02 |
| $\alpha$ decay scheme | Initial $\alpha(1 - t/1000)$ | - |
| Neighbourhood size | 5 | 1 |
| Neighbourhood function | Gaussian | - |
| Neighbourhood relation | Square | - |

Antigens are generated using system calls, captured through the use of strace and through manipulation within the antigen tissue client. The normalisation of the input signals is implemented using the tissue client, antigen processed using a separate tissue client, with data processing and the DCA performed using the tissue server process. An initial run of this system is performed to collect the input data and to check for any potential coding errors. Input signals and antigens are collected and recorded in a logfile using the real-time runs. Analysis of the preliminary real-time results of the output antigen and empirical analysis of the input data indicate its suitability for use in these experiments. The libtissue tcreplay client is used to perform the numerous runs of each data set. It is important to stress that the system is designed to work in real-time, though tcreplay is used to provide reproducibility of results, so a rigorous analysis can be performed.

5.6 DCA Results

The results of the DCA applied to the passive normal data are presented in Figures 17-19 and in Table 9 and 10. The results for an antigen segment size z=100 are shown in Figure 17 and represent results generated across ten runs by the DCA on the same data set. High MCAV values are shown upon the initiation of the SYN scan for both nmap and pts processes. The moving average trendline shows three distinct 'spikes' which correlate to the initial sending out of packets by the SYN scan, the period of targeted scanning and the final stages where network connections are terminated and scan results collated by the nmap process.

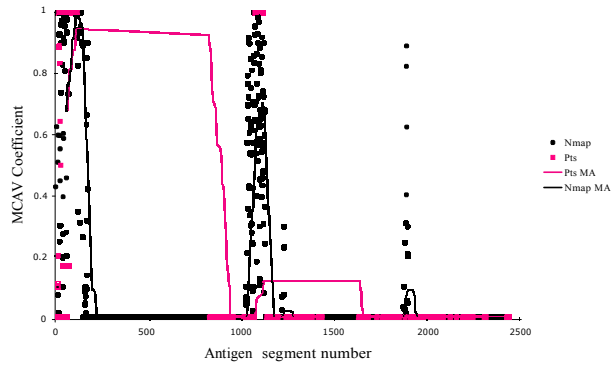

Fig. 17 PN results for the DCA for nmap SYN scan and pts processes. Points represent an average MCAV derived from across the ten runs performed, where the size of the antigen segment is z=100. The trendline represents a moving average across 50 data points.

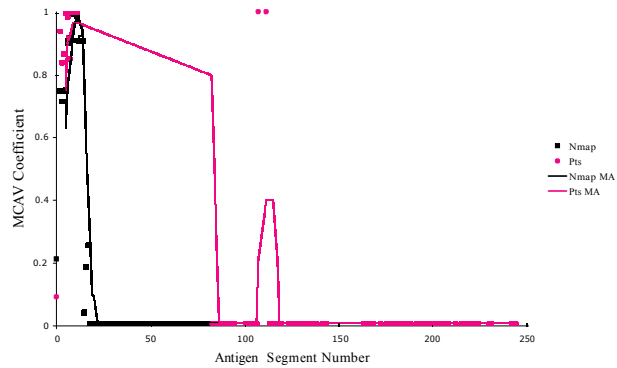

Fig. 18 PN results for the DCA for nmap SYN scan and pts processes The size of the antigen segments (z) is 1,000. This data represents an average MCAV derived from across the ten runs performed. The trendlines represent moving averages per process of interest across 20 data points.

Table 9 The results for both active (AN) and passive (PN) normal data sets for the DCA with an antigen segment size of z = 100,000. Segment 0 includes antigen 0-99,999; segment 1 includes antigen 100,000 to 199,999 etc.

| Segment Num | PN Nmap | PN Pts | AN Nmap | AN Firefox | AN Pts |
|---|---|---|---|---|---|
| 0 | 0.12 | 0.24 | 0.16 | 0.05 | 0.05 |
| 1 | 0.07 | 0.05 | 0.06 | 0.04 | 0.06 |
| 2 | 0 | 0 | 0.01 | 0.00 | 0.00 |
| 3 | - | - | 0.05 | 0.02 | 0.01 |

While the pts process produces a high MCAV initially, between antigen segments 100 and 900 no antigens are presented for the pts process, as it is inactive at this point. As the trendline is required to clarify the results this indicates that a higher value of z would be preferable to clearly assess the presence of an anomalous process.

In Figure 18, the result of the PN data set are presented where z=1,000. As with the results presented in Figure 17, an initial spike of a high MCAV is shown, implying that the scan is in its initial stages. While the individual points on this graph are not as dense as in Figure 17, the additional spikes representing the latter stages of the scan are less in magnitude, though little difference in the initial MCAV for the nmap processes is shown.

The results for z=10,000 are plotted in Figure 19 showing lower values for the pts process and a less sustained response to the nmap SYN scan process. Sensitivity is lost when the value of z is increased further as shown in Tables 9 and 10, with the maximum MCAVs greatly reduced from 1.0 in previous experiments to 0.19 for the nmap process.

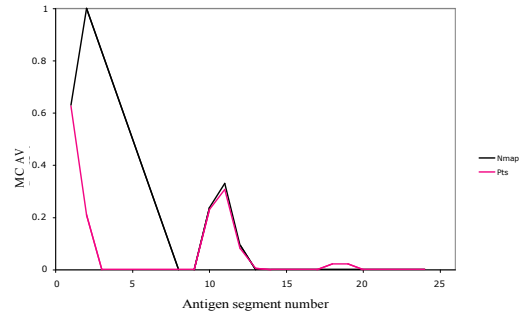

Fig. 19 PN results for the DCA for all three processes. Individual points not included for the sake of clarity. The size of the antigen segments (z) is 10,000. This data represents an average MCAV derived from across the ten runs performed. The trendlines represent moving averages per process of interest across ten data points.

Table 10 The results for the DCA for both active (AN) and passive (PN) normal data sets with an antigen segment size of z = 1,000,000.

| Process | AN | PN |
|---------|------|------|
| Nmap | 0.07 | 0.19 |
| Pts | 0.04 | 0 |
| Firefox | 0.03 | n/a |

The AN results produced by the DCA show similar features. Figure 20 shows the results using the smallest antigen segment size of z=100. In contrast to the PN data set, the nmap SYN scan is not invoked until antigen segment 500. As shown on this graph, following the initiation of this scan, the MCAV of all three processes of interest (nmap, pts, firefox) are shown to increase.

These increases in MCAV form four spikes throughout the session duration. The density of the datapoints in this graph makes the correct interpretation of this graph somewhat difficult. This is improved with the addition of a trendline, generated through applying a moving average of 50 points across the data. During the scan period, the DCA presents antigen, irrespective of its source, in the mature context, as shown through the generation of high MCAV values for both the nmap SYN scan and firefox browser processes. This implies that the generation of false positives occurs when a normal and anomalous process run simultaneously through the monitored ssh demon.

The results for antigen segments 0 to 500 are shown in Figure 21 for the sake of clarity. During this period the majority of antigen presented belong to the firefox process and some modulation of the behaviour of the monitored system occurs, as seen in the initial 500 seconds of Figure 20. Despite these activities, the MCAVs presented in Figure 21 are all relatively low. This suggests that the DCA using these particular signals responds appropriately to normal processes in the absence of scanning activity.

In Figure 22, the results are presented for the AN data set where z=1,000. In comparison to Figure 20, the trendlines of the graph are observably similar. Figure 23 shows that an antigen segment size of z=10,000 produces observably different results to z=1,000. This is evident as the major spike peak evident in Figure 20 and 22 is missing in Figure 23. Additionally, the only process technically classed as 'anomalous' (MCAV above 0.5, chosen to reflect the proportion of nmap antigen in the input data) is the nmap scan, though only briefly. This implies that the larger size of z increases the rate of false negatives as shown through the lower values in Figure 23 and also shown in Tables 9 and 10. For experiments where z = 100,000 and z=1,000,000, the MCAVs are also reduced. However, an interesting effect is that there is a greater difference in the MCAVs of the normal and anomalous processes.

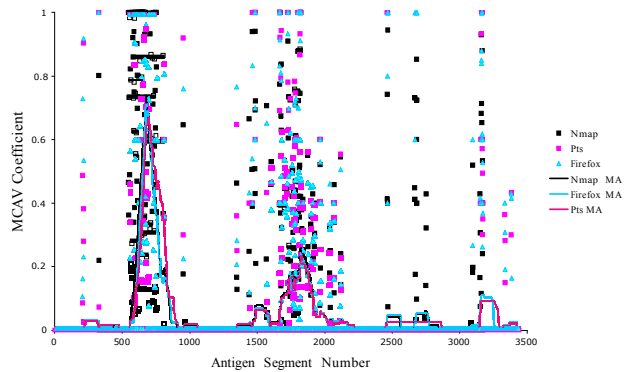

Fig. 20 AN results for the DCA for nmap SYN scan and pts processes. Points represent an average MCAV derived from across the ten runs performed, where the size of the antigen segment is z=100. The trendline represents a moving average across 50 data points.

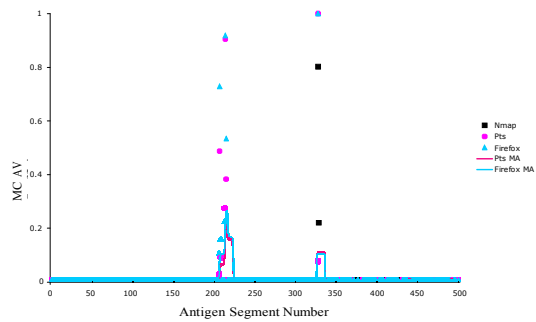

Fig. 21 AN results for initial 500 antigen segments produced with the DCA to highlight the low MCAVs yielded. Points represent an average MCAV derived from across the ten runs performed, where z = 100. The trendline represents a moving average across 20 data points.

An example of this is evident in antigen segment zero presented in Table 9, where the MCAV for the nmap scan is three times the magnitude of the MCAV of the firefox process. If the threshold of anomaly is applied at a level of 0.1, this experiment would yield results which detected the nmap SYN scan as anomalous and the firefox process as 'normal'. Other examples of this are also shown for the AN data set in Table 10. This implies that while larger values of z can produce false negatives, the potential for the reduction of false positives is also evident.

### 5.7 SOM Results

The PN results for the SOM are presented in Figures 24-26. As with the results of the DCA, high coefficient values are generated initially for both nmap and pts processes. A further 'spike' is evident at segment numbers 1,000-1,200 of Figure 24. These trends are also evident in Figures 25 and 26. Trendlines are added to each graph to represent a moving average per process. At the lowest level of granularity of z=1,800 (equivalent in the number of segments to DCA z=100), it is unclear as to exactly what the individual data points imply. Therefore a larger size of segment may be required, as also found with the DCA. Again, sensitivity is lost as the size of z is large, as shown by the results presented in Table 11 where z is 1,800,000.

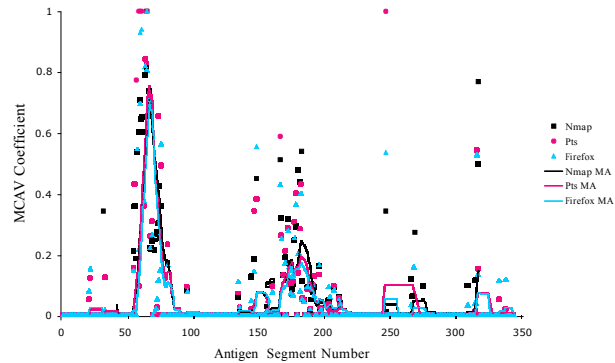

Fig. 22 AN results for all three processes produced with the DCA. The size of the antigen segments (z) is 1,000. This data represents an average MCAV derived from across the ten runs performed. The trendlines represent moving averages per process of interest across ten data points.

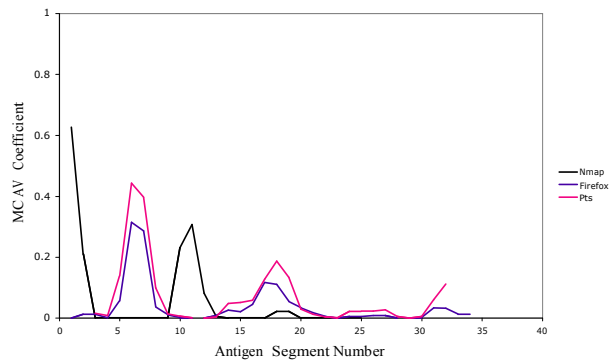

Fig. 23 AN results for all three processes. Individual points not included for the sake of clarity. The size of the antigen segments (z) is 10,000. This data represents an average MCAV derived from across the ten runs performed. The trendlines represent moving averages per process of interest across ten data points.

In a similar manner, the AN results for the SOM produce initially high coefficients for the nmap process. The results for $z=1,800$ are presented in Figure 27. This shows a major spike at the point of the scan commencement (segments 400-700). Unlike the DCA upon application of a trendline, it appears that the response to the scan is not sustained as three peaks are evident, as opposed to the single peak shown with the DCA. Also, the SOM produces high coefficient values for the firefox process, suggesting that discrimination between active anomalous and active normal processes can not be completely achieved by either algorithm.

The graphs produced for $z=18,000$ and $180,000$ are shown in Figures 28 and 29 respectively. As with the PN results, the response to the nmap decreases as the value of z increases. This is evident from both graphs and in Table 11. Unlike the DCA, which produced MCAVs for nmap which are consistently higher than with firefox (Figure 23), with the SOM results both the nmap and firefox coefficients decrease at a similar rate, as exemplified in Figure 29. Statistical analysis is presented to verify these observations in the next section.

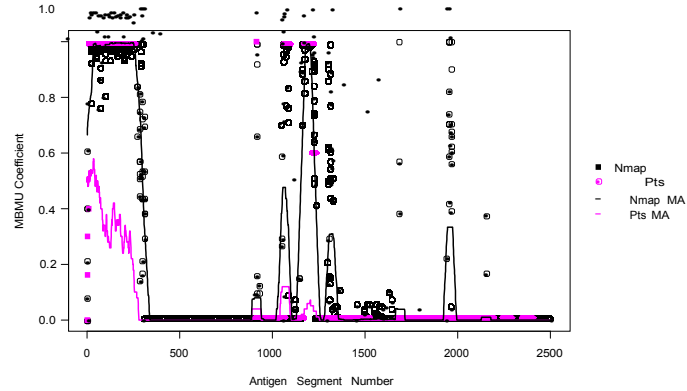

Fig. 24 PN results for all three processes using the SOM. The size of the output segments (z) is 1,800. This data represents an average MBMU derived from across the ten runs performed. The trendlines represent moving averages per process of interest across 50 data points.

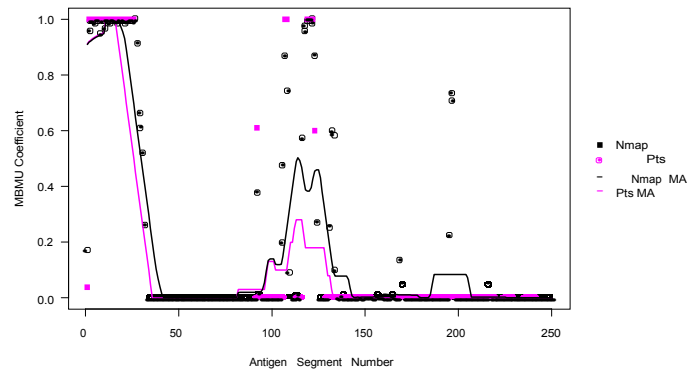

Fig. 25 PN results for all three processes using the SOM. The size of the output segments (z) is 18,000. This data represents an average MBMU derived from across the ten runs performed. The trendlines represent moving averages per process of interest across 20 data points.

Table 11 The results for both active (AN) and passive (PN) normal data sets for the SOM with a segment size of z = 1,800,000.

| Segment Num | PN Nmap | AN Nmap | AN Firefox |
|---|---|---|---|
| 0 | 0.293 | 0.078 | 0.028 |
| 1 | 0.135 | 0.016 | 0.013 |
| 2 | - | 0.003 | 0.011 |
| 3 | - | - | 0.005 |
| 4 | - | - | 0.002 |
| 5 | - | - | 0.001 |

## 6 Discussion: Analysis and Comparison

### 6.1 Null Hypothesis H1: Antigen segments

Performance of Mann-Whitney tests (an unpaired rank-based statistical test for non-parametric data [31]) comparing the results are presented in Figure 20 with the other antigen segment sizes. The results of this analysis are shown in Table 12, which assesses the rejection of null hypothesis H1. As indicated, in three out of the four tested cases, the data series are significantly different. This indicates that z does have an influence on the results of the DCA. Some further work with this concept may prove fruitful, especially if dynamic antigen segment sizes are used, linked to process activity. The demonstration of statistical significance implies that null hypothesis H1 can be rejected.

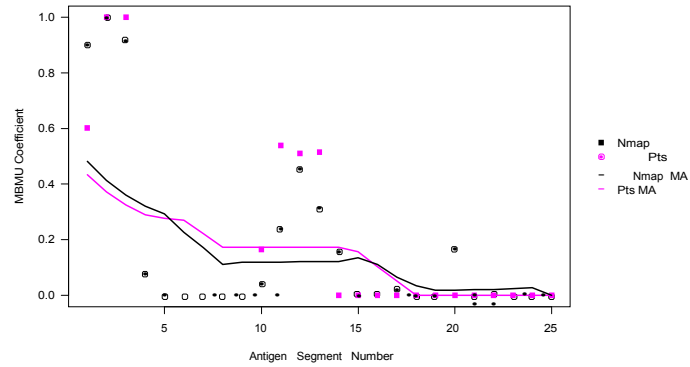

**Fig. 26** PN results for all three processes using the SOM. The size of the output segments (z) is 180,000. This data represents an average MBMU derived from across the ten runs performed. The trendlines represent moving averages per process of interest across ten data points.

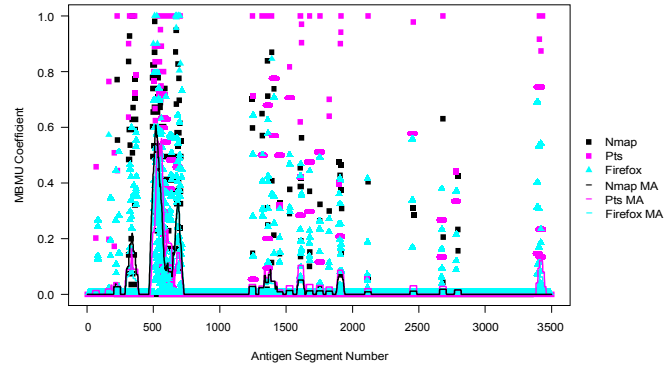

**Fig. 27** AN results for all three processes using the SOM. The size of the output segments (z) is 1,800. This data represents an average MBMU derived from across the ten runs performed. The trendlines represent moving averages per process of interest across 50 data points.

As the data is not normally distributed for either algorithm, Mann-Whitney tests are performed on the results of the SOM, comparing the results of z=1,800 with all other results for the AN nmap process. The resultant p-values are presented in Table 13. These results show that the modification of z produces a statistically significant effect on the resultant anomaly values. Therefore, null hypothesis H1 is also rejected for the SOM in addition to its rejection produced by the DCA.

## 6.2 Null Hypothesis H2: Head to Head Comparison

To assess H2, the results produced by the DCA and SOM are compared statistically for one selected antigen segment size. For this purpose DCA z=1,000 and SOM z=18,000 are used as they contain approximately a similar number of segments for both nmap and firefox. As these data are not normally distributed (confirmed by the application of the Shapiro-Wilk test) and do not form exact pairs, a one-sided unpaired Mann-Whitney test is used to perform this comparison. As the sample size is in excess of 300 datapoints, a 99% confidence interval is deemed appropriate for this assessment.

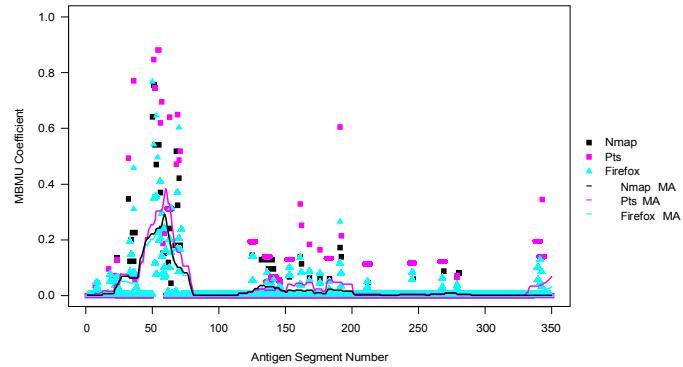

Fig. 28 AN results for all three processes using the SOM. The size of the output segments (z) is 18,000. This data represents an average MBMU derived from across the ten runs performed. The trendlines represent moving averages per process of interest across 20 data points.

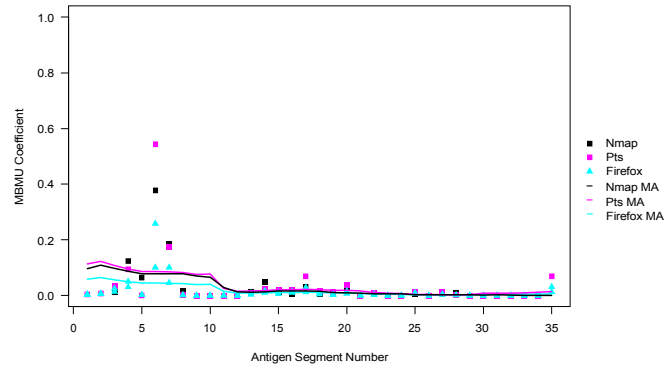

Fig. 29 AN results for all three processes using the SOM. The size of the output segments (z) is 180,000. This data represents an average MBMU derived from across the ten runs performed. The trendlines represent moving averages per process of interest across ten data points.

Table 12 The results of the Mann-Whitney test comparing the results of z=100 to the results of z=1,000, z=10,000 and z=100,000. A confidence interval of 95 % is used and data which are statistically significantly different are marked with an asterisk.

| $z$ | p-value |
|---|---|
| 1,000 | <0.0001* |
| 10,000 | 0.358 |
| 100,000 | < 0.0001* |

The results of this comparison for the firefox process yields a p-value of 0.02, which at the given confidence interval implies that the two sets of results are not statistically significant. This implies that the algorithms produce similar results for active normal processes. For the firefox process, null hypothesis H2 cannot be rejected under these particular circumstances with these given data sets.

Upon performance of the same statistical test, the nmap process produces a p-value of 0.002, which shows that the two algorithms produce statistically significant differences in the detection of the scan process. To assess which system produces the *better* performance, an additional two-sided Mann-

Table 13  The results of the Mann-Whitney test comparing the SOM results of z=1,800 to the results of z=18,000, z=180,000 and z=1,800,000. A confidence interval of 95 % is used and data which are statistically significantly different are marked with an asterisk.

| $z$ | p-value |
|---|---|
| 18,000 | <0.0001* |
| 180,000 | < 0.0001* |
| 1,800,000 | < 0.0001* |

Whitney is performed. The results of this test show that the DCA has the improved performance, producing a p-value of 0.0001. Therefore null hypothesis H2 can be rejected for the nmap process and that the DCA shows the better performance on this occasion.

### 6.3 Baseline

To validate both sets of results and to ensure that both performances are improved over a baseline, a k-means classifier is applied to the signal data. The classifier used belongs to the WEKA suite [65]. In this test 52% of the signals were classed as belonging to one class while 48% to another class. This implies that the necessary discrimination cannot be achieved through classification on the basis of signals alone. This also shows that this data is non-trivial to classify and adds value to the results produced for both the SOM and DCA.

### 6.4 Conjecture

We have validated the use of the DCA as a serious competitor for anomaly detection applications. Until this comparison we were still uncertain as to the quality of results produced by the DCA. This comparison with the traditional SOM has shown that the DCA shows great promise as a successful AIS algorithm. The performance produced by the DCA shows that the algorithm is capable of performing at a level comparable to a standard technique.

It is interesting that the results for the firefox process are not statistically different for both algorithms, yet differences are evident for the nmap process. This may be attributed to the method by which the signals are processed. In the DCA, signals are assigned weights on a per-category basis. In the SOM, all signals carry equal weight. Perhaps the fact that the DCA uses heavy weighting for both PAMP and danger signals is responsible for the improved rate of nmap detection.

Alternatively, the correlation between antigen and signal occurs within the DCA but is performed explicitly during the normalisation stage with the SOM. This activity-dependent association produced by the DCA may mean that for the duration of the scan, the coupling between the signals and antigen for the nmap process is tighter. This could be due to the fact that the amount of processed antigen is increased. As noted by Oates *et al* [57] the increased DC migration rate may result in greater volumes of processed antigen. This may result in a tighter coupling between antigen and signals during periods of high activity and high signal levels. A theoretical analysis, outside of the scope of this paper, is necessary to confirm the coupling mechanism and may provide insight as to the cause and nature of this effect plus the further reaching consequences of applicability of the DCA.

### 7 Conclusions

In this paper we have compared two biologically inspired algorithms, the immune-inspired DCA and the neurological-inspired SOM when applied to the detection of a SYN port scan. Two constructed data sets are produced for this purpose consisting of 13 million data items to classify. Each algorithm was successful at performing anomaly detection, with the number of antigen, z, used per classification step varied. It is shown that both algorithms are sensitive to changes in z. A direct comparison between the SOM and DCA is provided. For the normal processes, both algorithms performed equally well, with the DCA producing a significantly improved performance at detecting the anomalous process. As a result, the DCA is shown to be a competitive anomaly detection algorithm. As further work, the DCA will be applied experimentally to other large data sets, including further benchmarking with computer security data and applications potentially including large-scale data which requires correlation, such as data derived from radio telescopes. In addition a theoretical analysis of the DCA is required to fully understand the nature of this relatively novel algorithm and to be able to assess future successful applications.

## References

1. U. Aickelin, P. Bentley, S. Cayzer, J. Kim, and J. McLeod. Danger theory: The link between AIS and IDS. In *Proc. of the 2nd International Conference on Artificial Immune Systems (ICARIS), LNCS 2787*, pages 147–155. Springer-Verlag, 2003.
2. S. Albayrak, C. Scheel, D. Milosevic, and A. Muller. Combining Self-Organizing Map Algorithms for Robust and Scalable Intrusion Detection. In *International Conference on Computational Intelligence for Modelling, Control and Automation and International Conference on Intelligent Agents, Web Technologies and Internet Commerce*, volume 2, 2005.
3. M. Amini, R. Jalili, and H.R. Shahriari. RT-UNNID: A practical solution to real-time network-based intrusion detection using unsupervised neural networks. *Computers & Security*, 25(6):459–468, 2006.
4. C. Bailey-Lee, C. Roedel, and E. Silenok. Detection and characterization of port scan attacks. Technical report, University of California San Diego (UCSD), 2003.
5. J. Balthrop, F. Esponda, S. Forrest, and M. Glickman. Coverage and generaliszation in an artificial immune system. In *Proc. of the Genetic and Evolutionary Computation Conference (GECCO)*, pages 3–10, 2002.
6. R. Bejtlich. *Extrusion Detection: Security Monitoring for Internal Intrusions*. Addison-Wesley Professional, 2005.
7. P. Bentley, J. Greensmith, and S. Ujjin. Two ways to grow tissue for artificial immune systems. In *Proc. of the 4th International Conference on Artificial Immune Systems (ICARIS), LNCS 3627*, pages 139–152. Springer-Verlag, 2005.
8. A. Bivens, C. Palagiri, R. Smith, B. Szymanski, and M. Embrechts. Network-Based Intrusion Detection Using Neural Networks. *Intelligent Engineering Systems through Artificial Neural Networks*, 12(1):579–584, 2002.
9. D. Bolzoni, S. Etalle, P. Hartel, and E. Zambon. Poseidon: a 2-tier anomaly-based network intrusion detection system. In *Fourth IEEE International Workshop on Information Assurance (IWIA '06)*, volume 0, pages 144–156, Los Alamitos, CA, USA, 2006. IEEE Computer Society.
10. Sung-Bae Cho. Incorporating soft computing techniques into a probabilistic intrusion detection system. *IEEE Transactions on Systems, Man and Cybernetics*, 32(2):154–160, May 2002.
11. J. Choy and S.B. Cho. Anomaly Detection of Computer Usage Using Artificial Intelligence Techniques. *Advances in Artificial Intelligence: PRICAI 2000*, 2112:31–43, 2001.
12. R. Coico, G. Sunshine, and E. Benjamini. *Immunology: A Short Course*. Wiley-Liss, 2003.
13. S. Cross, R. Harrison, and R. Kennedy. Introduction to neural networks. *The Lancet*, 346(8982):1075–1079, October 1995.
14. L. de Castro and J. Timmis. *Artificial Immune Systems: A New Computational Approach*. Springer-Verlag, London. UK., September 2002.
15. L DeLooze. Attack Characterization and Intrusion Detection using an Ensemble of Self-Organizing Maps. In *International Joint Conference on Neural Networks (IJCNN '06)*, pages 2121–2128, 2006.
16. O. Depren, M. Topallar, E. Anarim, and M.K. Ciliz. An intelligent intrusion detection system (IDS) for anomaly and misuse detection in computer networks. *Expert Systems With Applications*, 29(4):713–722, 2005.
17. F. Dostoevsky. nmap. http://www.insecure.org, last accessed, 5/10/07.
18. S. Forrest, A. Perelson, L. Allen, and R. Cherukuri. Self-nonself discrimination in a computer. In *Proc. of the IEEE Symposium on Security and Privacy*, pages 202–209. IEEE Computer Society, 1994.
19. K.L. Fox, R.R. Henning, J.H. Reed, and R. Simonian. A neural network approach towards intrusion detection. *Proceedings of the 13th National Computer Security Conference*, 10, 1990.
20. D. Gollmann. *Computer Security*. John Wiley & Sons, Inc. New York, NY, USA, 1999.
21. F. Gonzalez and D. Dasgupta. Neuro-immune and self-organizing map approaches to anomaly detection: A comparison. In *Proceedings of the 1st International Conference on Artificial Immune Systems*, pages 203–211, 2002.
22. F. Gonzalez and D. Dasgupta. Anomaly detection using real-valued negative selection. *Journal of Genetic Programming and Evolvable Machines*, 4:383–403, 2003.
23. F.A. González, J.C. Galeano, D.A. Rojas, and A. Veloza-Suan. Discriminating and visualizing anomalies using negative selection and self-organizing maps. In *Proceedings of the 2005 Conference on Genetic and Evolutionary Computation*, pages 297–304. ACM Press New York, NY, USA, 2005.
24. J. Greensmith. *The Dendritic Cell Algorithm*. PhD thesis, School of Computer Science, University Of Nottingham, 2007.
25. J. Greensmith and U. Aickelin. Dendritic cells for syn scan detection. In *Proceedings of the Genetic and Evolutionary Computation Conference (GECCO 2007)*, pages 49–56, 2007.

26. J. Greensmith, U. Aickelin, and S. Cayzer. Introducing Dendritic Cells as a novel immune-inspired algorithm for anomaly detection. In *Proc. of the 4th International Conference on Artificial Immune Systems (ICARIS), LNCS 3627*, pages 153–167. Springer-Verlag, 2005.

27. J. Greensmith, U. Aickelin, and G. Tedesco. Information fusion for anomaly detection with the dca. *Information Fusion*, tbc(tbc):tbc, 2008.

28. J. Greensmith, U. Aickelin, and J. Twycross. Articulation and clarification of the dendritic cell algorithm. In *Proc. of the 5th International Conference on Artificial Immune Systems (ICARIS), LNCS 4163*, pages 404–417, 2006.

29. J. Greensmith, J. Twycross, and U. Aickelin. Dendritic cells for anomaly detection. In *Proc. of the Congress on Evolutionary Computation (CEC)*, pages 664–671, 2006.

30. H. Gunes Kayacik, A. Nur Zincir-Heywood, and M.I. Heywood. A hierarchical SOM-based intrusion detection system. *Engineering Applications of Artificial Intelligence*, 20(4):439–451, 2007.

31. J.J. Higgins. *An Introduction to Modern Nonparametric Statistics*. Thomson Learning, 2004.

32. S. Hofmeyr and S. Forrest. Intrusion detection using sequences of system calls. *Journal of Computer Security*, 6:151–180, 1998.

33. S. Hofmeyr and S. Forrest. Immunity by design. In *Proc. of the Genetic and Evolutionary Computation Conference (GECCO)*, pages 1289–1296, 1999.

34. A. Höglund and K. Hätönen. Computer network user behaviour visualization using self organizing maps. In L. Niklasson, M. Bodén, and T. Ziemke, editors, *Proceedings of ICANN98, the 8th International Conference on Artificial Neural Networks*, volume 2, pages 899–904. Springer, London, 1998.

35. A. Höglund, K. Hätönen, and A. Sorvari. A computer host-based user anomaly detection system using the self-organizing map. In *IJCNN (5)*, pages 411–416, 2000.

36. T. Horeis. Intrusion Detection with Neural Networks–Combination of Self-Organizing Maps and Radial Basis Function Networks for Human Expert Integration. *Computational Intelligence Society Student Research Grants*, 2003.

37. Z. Ji and D. Dasgupta. Real-valued negative selection algorithm with variable-sized detectors. In *Proc. of the Genetic and Evolutionary Computation Conference (GECCO)*, pages 287–298, 2004.

38. C. Jirapummin, N. Wattanapongsakorn, and P. Kanthamanon. Hybrid neural networks for intrusion detection system. In *2002 International Technical Conference on Circuits/Systems,Computers and Communications (ITC-CSCC 2002)*, pages 928–931, Phuket, Thailand, 2002.

39. J. Jung, V. Paxson, A. Berger, and H. Balakrishnan. Fast portscan detection using sequential hypothesis testing. *Security and Privacy, IEEE Symposium on*, pages 211–225, 2004.

40. E.R. Kandel, J.H. Schwartz, and T.M. Jessell. *Principles of Neural Science*. McGraw-Hill/Appleton & Lange, 2000.

41. H. Kayacik, A. Zincir-Heywood, and M Heywood. On the capability of an SOM based intrusion detection system. In *Proceedings of the International Joint Conference on Neural Networks*, volume 3, 2003.

42. H.G. Kayacik and N. Zincir-Heywood. Analysis of Three Intrusion Detection System Benchmark Datasets Using Machine Learning Algorithms. In *Proceedings of IEEE International Conference on Intelligence and Security Informatics (ISI 2005)*, volume 3495 of *LNCS*, pages 362–367, Atlanta, GA, USA, May 19-20 2005. Springer.

43. R. Khanna and H. Liu. System approach to intrusion detection using hidden Markov model. *International Conference On Communications And Mobile Computing*, pages 349–354, 2006.

44. J. Kim and P. Bentley. Evaluating negative selection in an artificial immune system for network intrusion detection. In *Proc. of the Genetic and Evolutionary Computation Conference (GECCO)*, pages 1330 – 1337, July 2001.

45. J. Kim, P. Bentley, U. Aickelin, J. Greensmith, G. Tedesco, and J. Twycross. Immune system approaches to intrusion detection - a review. *Natural Computing*, page tbc, 2007. to appear, accepted for publication.

46. T. Kohonen. Automatic formation of topological maps of patterns in a self-organizing system. *Proceedings of the 2nd Scandinavian Conference on Image Analysis*, pages 214–220, 1981.

47. T. Kohonen. *Self-Organizing Maps*. Springer-Verlag, Berlin, 1996.

48. T. Kohonen. The self-organizing map. *Proceedings of the IEEE*, 78(9):1464–1480, Sep 1990.

49. SC Lee and DV Heinbuch. Training a neural-network based intrusion detector to recognizenovel attacks. *Systems, Man and Cybernetics, Part A, IEEE Transactions on*, 31(4):294–299, 2001.

50. JZ Lei and A. Ghorbani. Network intrusion detection using an improved competitive learning neural network. In *Second Annual Conference on Communication Networks and Services Research*, pages 190–197, 2004.

51. P. Lichodzijewski, A. Nur Zincir-Heywood, and M. Heywood. Dynamic intrusion detection using self organizing maps. In *The 14th Annual Canadian Information Technology Security Symposium (CITSS)*, 2002.

52. M. Lutz and G. Schuler. Immature, semi-mature and fully mature dendritic cells: which signals induce tolerance or immunity? *Trends in Immunology*, 23(9):991–1045, 2002.

53. P. Matzinger. Tolerance, danger and the extended family. *Annual Reviews in Immunology*, 12:991–1045, 1994.

54. P. Matzinger. Friendly and dangerous signals: is the tissue in control? *Nature Immunology*, 8(1):11–13, 2007.

55. P. Miller and A. Inoue. Collaborative intrusion detection system. In *22nd International Conference of the North American Fuzzy Information Processing Society (NAFIPS 2003)*, pages 519–524, 2003.

56. K. Murphy, P. Travers, and M. Walport. *Janeway's Immunobiology*. Garland Science, 7th edition, 2008.

57. R. Oates, J. Greensmith, U. Aickelin, J. Garibaldi, and G. Kendall. The application of a dendritic cell algorithm to a robotic classifier. In *Proc. of the 6th International Conference on Artificial Immune Systems (ICARIS), LNCS 4628*, pages 204–215, 2007.

58. R. Oates, G. Kendall, and J. Garibaldi. Frequency analysis for dendritic cell population tuning: Decimating the dendritic cell. *Submitted to Evolutionary Intellegence*, 2007.

59. M. Ramadas, S. Ostermann, and B. Tjaden. Detecting Anomalous Network Traffic with Self-organizing Maps. In *Proceedings of Recent Advances in Intrusion Detection: 6th International Symposium (RAID 2003)*, Pittsburgh, Pa, Usa, September 8-10 2003. Springer.

60. B. C. Rhodes, J. A. Mahaffey, and J. D. Cannady. Multiple self-organizing maps for intrusion detection. In *Proceedings of the 23rd National Information Systems Security Conference*, 2000.

61. H. Ritter, T. Martinetz, and K. Schulten. *Neural Computation and Self-Organizing Maps; An Introduction*. Addison-Wesley Longman Publishing Co., Inc. Boston, MA, USA, 1992.

62. M. Roesch. Snort - lightweight intrusion detection for networks. In *Proc. of the 13th USENIX conference on System administration (LISA)*, pages 229–238. USENIX Association, 1999.

63. S. Sarasamma and Q. Zhu. Min-max hyperellipsoidal clustering for anomaly detection in network security. *Systems, Man and Cybernetics, Part B, IEEE Transactions on*, 36(4):887–901, 2006.

64. S.T. Sarasamma, Q.A. Zhu, and J. Huff. Hierarchical Kohonenen net for anomaly detection in network security. *IEEE Transactions on Systems, Man, and Cybernetics, Part B: Cybernetics*, 35(2):302–312, 2005.

65. S.Garner. Weka: The waikato environment for knowledge analysis. In *In Proc. of the New Zealand Computer Science Research Students Conference*, pages 57–64, 1995.

66. A. Somayaji, M. Locasto, and J. Feyereisl. Panel on the future of biologically-inspired security: Is there anything left to learn? In *New Security Paradigms Workshop (NSPW'07)*, 2007.

67. R. Sporri and C. Caetano. Inflammatory mediators are insufficient for full dendritic cell activation and promote expansion of cd4+ t cell populations lacking helper function. *Nature Immunology*, 6(2):163–170, January 2005.

68. S. Staniford, J. Hoagland, and J. McAlerney. Practical automated detection of stealthy portscans. *Journal of Computer Security*, 10(1-2):105–136, 2002.

69. T. Stibor, P. Mohr, J. Timmis, and C. Eckert. Is negative selection appropriate for anomaly detection? In *Proc. of Genetic and Evolutionary Computation Conference (GECCO)*, pages 321–328, 2005.

70. T. Stibor, J. Timmis, and C. Eckert. Artificial immune systems for IT-security. *it- Information technology*, 48(3), 2006.

71. T. Stibor, J. Timmis, and C. Eckert. On permutation masks in hamming negative selection. In *Proc. of the 5th International Conference on Artificial Immune Systems (ICARIS), LNCS 4163*, pages 122–135, 2006.

72. J. Timmis. Artificial immune systems: today and tomorrow. *Natural Computing*, 6(1):1–18, March 2007.

73. J. Twycross. *Integrated Innate and Adaptive Artificial Immune Systems Applied to Process Anomaly Detection*. PhD thesis, University Of Nottingham, 2007.

74. J. Twycross and U. Aickelin. Towards a conceptual framework for innate immunity. In *Proc. of the 4th International Conference on Artificial Immune Systems (ICARIS), LNCS 3627*, pages 112–125. Springer-Verlag, 2005.

75. J. Twycross and U. Aickelin. libtissue - implementing innate immunity. In *Proc. of the Congress on Evolutionary Computation (CEC)*, pages 499–506, 2006.

76. W. Wang, X. Guan, X. Zhang, and L. Yang. Profiling program behavior for anomaly intrusion detection based on the transition and frequency property of computer audit data. *Computers & Security*, 25(7):539–550, 2006.

77. C. Williams, R. Harry, and J. McLeod. Mechanisms of apoptosis induced DC suppression. *Submitted to the Journal of Immunology*, 2007.

78. I.S. Yoo. Visualizing windows executable viruses using self-organizing maps. In *Proceedings of the 2004 ACM workshop on Visualization and data mining for computer security*, pages 82–89. ACM Press New York, NY, USA, 2004.